\documentclass{article}

\PassOptionsToPackage{numbers}{natbib}

\usepackage[final]{neurips_2024}

\usepackage[utf8]{inputenc} %
\usepackage[T1]{fontenc}    %
\usepackage{hyperref}       %
\usepackage{url}            %
\usepackage{booktabs}       %
\usepackage{amsfonts}       %
\usepackage{nicefrac}       %
\usepackage{microtype}      %
\usepackage{xcolor}         %

\usepackage{ulem}
\usepackage{enumitem}
\setlist[itemize]{leftmargin=10mm}
\usepackage{graphicx}
\usepackage{amsmath}
\usepackage{amssymb}
\usepackage{booktabs}
\usepackage{caption}
\usepackage{multirow}
\usepackage{bbm}
\usepackage{wrapfig}
\usepackage{makecell}
\usepackage{bbding}
\usepackage{pifont}
\usepackage{wasysym}
\usepackage{amssymb}
\usepackage{color, colortbl}
\usepackage{mathtools}
\usepackage{lipsum}
\usepackage{arydshln}
\usepackage{comment}

\usepackage{algorithm}
\usepackage{algpseudocode}
\algrenewcommand\algorithmicfunction{\textbf{Function}}

\hypersetup{
    colorlinks=true,
    filecolor=magenta,      
    urlcolor=black,
    citecolor=cyan,
    pdfpagemode=FullScreen,
}

\newcommand{\tabincell}[2]{\begin{tabular}{@{}#1@{}}#2\end{tabular}} 
\newcommand{\hr}[1]{\textcolor{black}{#1}}
\newcommand{\hry}[1]{\textcolor{black}{#1}}

\usepackage{array}
\newcolumntype{x}[1]{>{\centering\arraybackslash\hspace{0pt}}p{#1}}
\newcommand{\niparagraph}[1]{\vspace{2pt}\noindent\textbf{#1}}
\definecolor{maroon}{cmyk}{0,0.87,0.68,0.32}
\definecolor{n1}{HTML}{ff9999}
\definecolor{n2}{HTML}{FFCC99}
\definecolor{n3}{HTML}{FFFF99}

\title{ShiftAddLLM: Accelerating Pretrained LLMs via Post-Training Multiplication-Less Reparameterization}

\usepackage{authblk}
\usepackage{xpatch}

\xpatchcmd{\author}{\relax#1\relax}{\relax\detokenize{#1}\relax}{}{}
\author[\empty]{\textbf{Haoran You}\textsuperscript{$\dagger$}, \textbf{Yipin Guo}\textsuperscript{$\dagger$}, \textbf{Yichao Fu}\textsuperscript{$\dagger$}, \textbf{Wei Zhou}\textsuperscript{$\dagger$}, \textbf{Huihong Shi}\textsuperscript{$\dagger$}, \textbf{Xiaofan Zhang}\textsuperscript{$*$}}
\author[\empty]{\\ \textbf{Souvik Kundu}\textsuperscript{$\diamond$}, \textbf{Amir Yazdanbakhsh}\textsuperscript{$\ddagger$}, \textbf{Yingyan (Celine) Lin}\textsuperscript{$\dagger$}}

\xpatchcmd{\affil}{\relax#1\relax}{\relax\detokenize{#1}\relax}{}{}
\affil[$\dagger$]{Georgia Institute of Technology \quad $^\diamond$Intel Labs \quad $^*$Google \quad $^\ddagger$Google DeepMind}

\affil[$\dagger$]{\textit{\{haoran.you, celine.lin\}@gatech.edu, eic-lab@groups.gatech.edu}}
\affil[$^\diamond$]{\textit{souvikk.kundu@intel.com, $^{*\ddagger}$\{xiaofanz, ayazdan\}@google.com}}

\begin{document}

\maketitle

\begin{abstract}
Large language models (LLMs) have shown impressive performance on language tasks but face challenges when deployed on resource-constrained devices due to their extensive parameters and reliance on dense multiplications, resulting in high memory demands and latency bottlenecks. 
Shift-and-add reparameterization offers a promising solution by replacing costly multiplications with hardware-friendly primitives in both the attention and multi-layer perceptron (MLP) layers of an LLM. However, current reparameterization techniques require training from scratch or full parameter fine-tuning to restore accuracy, which is resource-intensive for LLMs. 
To address this, we propose accelerating pretrained LLMs through post-training shift-and-add reparameterization, creating efficient multiplication-free models, dubbed ShiftAddLLM. 
Specifically, we quantize each weight matrix into binary matrices paired with group-wise scaling factors. The associated multiplications are reparameterized into (1) shifts between activations and scaling factors and (2) queries and adds according to the binary matrices.
To reduce accuracy loss, we present a multi-objective optimization method to minimize both weight and output activation reparameterization errors. Additionally, based on varying sensitivity across layers to reparameterization, we develop an automated bit allocation strategy to further reduce memory usage and latency.
Experiments on five LLM families and eight tasks consistently validate the effectiveness of ShiftAddLLM, achieving average perplexity \hry{reductions} of 5.6 and 22.7 points at comparable or lower latency compared to the most competitive quantized LLMs at 3- and 2-bit precision, respectively, and more than 80\% memory and energy reductions over the original LLMs. Codes and models are available at \url{https://github.com/GATECH-EIC/ShiftAddLLM}.
\end{abstract}

\section{Introduction}
Pretrained LLMs have demonstrated state-of-the-art performance in language understanding and generation tasks~\cite{chatgpt,gpt4,bard,team2023gemini,zhang2022opt,touvron2023llama,touvron2023llama2,llama3}. 
However, deploying these LLMs incurs significant hardware demands, including high latency, memory, and energy consumption, especially on edge or cloud GPU devices.
The primary bottlenecks are their immense parameter sizes and the associated multiplication operations. For instance, GPT-3, with 175 billion parameters, requires 350GB of memory in FP16 format~\cite{lin2023awq} and performs $10^{15}$ floating-point operations (FLOPs) for a single forward pass~\cite{flops_memory_summary}. Previous efforts to improve LLM efficiency have focused on pruning~\cite{ma2023llm,sun2023simple,gromov2024unreasonable,qs-theory,slope}, quantization~\cite{xiao2023smoothquant,lin2023awq,frantar2022optq,park2022lut}, and attention optimization~\cite{dao2022flashattention,you2024linear,yang2023gated}. 
However, these methods still rely on costly multiplication operations in both the attention and MLP layers.

We identify a promising yet unexplored opportunity for improving LLM efficiency: \textit{reparameterizing their extensive multiplications with more cost-effective hardware substitutes, such as bitwise shifts and adds.}
Inspired by practices in computer architecture and digital signal processing, replacing multiplications with bitwise shifts and adds~\cite{xue1986adaptive,gwee2008low} can offer up to \hry{$\nicefrac{3.1}{0.1} =$ 31$\times$ energy and $\nicefrac{3495}{137} \approx$ 26$\times$ area reductions} (see Tab. \ref{tab:unit}). This hardware-inspired approach can lead to efficient and fast implementations, as shown by previous research on ShiftAddNet~\cite{you2020shiftaddnet,you2022shiftaddnas,you2024shiftaddvit}. Unlike previous techniques that require training from scratch or extensive fine-tuning, we propose a new method to integrate the shift-and-add concept into LLMs through post-training optimization.
\begin{table}[t]
\centering
\renewcommand{\arraystretch}{1.2}
\caption{Hardware cost under 45nm CMOS~\cite{horowitz20141,you2020shiftaddnet,han2016eie,sentieys2021approximate,brito2014quaternary}.}
\resizebox{0.95\linewidth}{!}{
\begin{tabular}{c|cccc|cccc|ccc|c}
\Xhline{3\arrayrulewidth}
\multirow{2}{*}{\textbf{OPs}} & \multicolumn{4}{c|}{\textbf{Multiplication}} & \multicolumn{4}{c|}{\textbf{Add}} & \multicolumn{3}{c|}{\textbf{Shift}} & \multirow{2}{*}{\textbf{\tabincell{c}{LUTs\\(8-bit Query)}}} \\
\cline{2-5} \cline{6-9} \cline{10-12} %
 & \texttt{FP32} & \texttt{FP16} & \texttt{INT32} & \texttt{INT8} & \texttt{FP32} & \texttt{FP16} & \texttt{INT32} & \texttt{INT8} & \texttt{INT32} & \texttt{INT16} & \texttt{INT8} & \\ %
\Xhline{3\arrayrulewidth}
\textbf{Energy (pJ)} & 3.7 & 0.9 & 3.1 & 0.2 & 1.1 & 0.4 & 0.1 & 0.03 & 0.13 & 0.057 & 0.024 & 0.37 (8 OPs) \\
\textbf{Area ($\mu$m$^2$)} & 7700 & 1640 & 3495 & 282 & 4184 & 1360 & 137 & 36 & 157 & 73 & 34 & 787 (8 OPs) \\
\Xhline{3\arrayrulewidth}
\end{tabular}
}
\label{tab:unit}
\leftline{\small \quad\quad * Note that 1 LUT corresponds to 8 operations, as each bit in queries is from a weight element.}
\end{table}

To design multiplication-less LLMs, we need to address three key challenges:
\textbf{\textit{First}}, how can we effectively reparameterize pretrained LLMs with shifts and adds in a post-training manner? Previous reparameterization techniques~\cite{you2020shiftaddnet,you2024shiftaddvit} can result in nontrivial quantization errors, requiring fine-tuning or retraining to avoid accuracy drops. 
We aim to develop a ready-to-use \textit{post-training} reparameterization method for LLMs.
\textbf{\textit{Second}}, how can we mitigate the accuracy drop from shift-and-add reparameterization? Approximating original multiplications with lower-bit shifts and adds typically reduces model accuracy. Most studies resort to fine-tuning or increasing model sizes, complicating LLM deployment. We hypothesize that optimizing both weight and activation errors can minimize overall reparameterization error, aligning with recent activation-aware weight quantization methods in LLMs.
\textbf{\textit{Third}}, how can we handle varying sensitivities to reparameterization across different layers and blocks in LLMs? An automated strategy to determine the optimal number of bits for reparameterized weights in each layer is needed. More vulnerable layers should have higher-bit representations, while less sensitive layers can use lower-bit representations. This ensures no bottlenecked layers due to aggressive reparameterization and maximizes redundancy exploitation.
To the best of our knowledge, \textit{this is the first attempt} to address these three challenges for multiplication-less LLMs through \textit{post-training} reparameterization.
Our contributions are summarized as follows:
\begin{itemize}[leftmargin=*]
    \item We propose accelerating pretrained LLMs via a \textit{post-training} bitwise shift-and-add reparameterization, resulting in efficient multiplication-less LLMs, dubbed ShiftAddLLM. All weights are quantized into binary matrices paired with group-wise scaling factors; the associated multiplications are reparameterized into shift-and-add operations.
    \item To mitigate accuracy loss, we present a multi-objective optimization method aligning and optimizing both weight and output activation objectives, minimizing overall reparameterization error, and achieving lower perplexity and better task accuracy.
    \item We introduce a mixed and automated bit allocation strategy that determines the optimal number of bits for reparameterized weights per layer, based on their vulnerability to compression. Susceptible layers receive higher-bit representations, while less sensitive ones get lower-bit representations.
\end{itemize}
Our extensive results across five LLMs and eight tasks consistently show the superior accuracy and efficiency trade-offs achieved by ShiftAddLLM,
with average perplexity \hry{reductions} of 5.6 and 22.7 at comparable or even lower latency compared to the most competitive quantized LLMs at three and two bits, respectively, and more than 80\% memory and energy reductions over the original LLMs.

\section{Related Works}

\niparagraph{LLM Quantization.}
Significant efforts have been made to quantize LLMs, including quantization-aware training (QAT)~\cite{liu2023llm,shen2024edgeqat} and post-training quantization (PTQ)~\cite{frantar2022optq,lin2023awq,xiao2023smoothquant,dettmers2022gpt3}.
QAT requires calibrated data and significant retraining resources, whereas PTQ is more dominant due to it lower computational and time overhead.
There are two prevalent PTQ strategies for LLMs:
\textbf{\textit{(1)}} uniform quantization of both weights and activations~\cite{xiao2023smoothquant,dettmers2022gpt3,yao2022zeroquant}, often limited to 8 bits (W8A8) as lower bit representations can significantly reduce accuracy;
and \textbf{\textit{(2)}} lower bit weight-only quantization~\cite{frantar2022optq,park2022lut,dettmers2023case,huang2024billm,chee2024quip}, which quantizes LLM weights to lower bits while keeping activations in a \texttt{FP16} format. This approach alleviates memory bottlenecks associated with the vast parameters of LLMs. For instance, GPTQ~\cite{frantar2022optq} uses gradient-based weight quantization and develops \texttt{INT3/4} kernels to reduce data movements, and LUT-GEMM~\cite{park2022lut} eliminates the dequantization and uses custom LUT-based CUDA kernels to reduce memory and computation costs.
In contrast, ShiftAddLLM is the first to employ the shift-and-add idea for reparameterizing pre-trained LLMs. This reparameterization reduces bit usage for weights and replaces costly multiplications with hardware-friendly primitives,  further reducing energy,
latency, and memory.

\niparagraph{Multiplication-less Models.}
The efficient model community has focused on reducing or replacing multiplications.
In CNNs, binary networks~\cite{courbariaux2016binarized,juefei2017local} binarize weights and activations, while shift-based networks use spatial shifts~\cite{wu2018shift} or bitwise shifts~\cite{elhoushi2021deepshift} to substitute for multiplications. AdderNet~\cite{chen2020addernet, adder_distillation, adder_hardware} replaces multiplications with additions, albeit with a small accuracy drop. ShiftAddNet~\cite{you2020shiftaddnet} reparameterizes CNNs with cascaded shift and add layers.
These techniques have been adapted to Transformers.
BiLLM~\cite{huang2024billm} introduces binary LLMs, while \cite{shu2021adder} and \cite{wang2022shift} extend the addition or shift concepts to the attention mechanisms, respectively.
ShiftAddViT~\cite{you2024shiftaddvit} reparameterizes pretrained Vision Transformers (ViTs) with shifts and adds. 
Contemporary work MatMul-free LM~\cite{zhu2024scalable} leverages additive operators and Hadamard products for multiplication-free language model training, relying on FPGAs for speedups.
Compared to closely related works like ShiftAddNet~\cite{you2020shiftaddnet} and MatMul-free LM~\cite{zhu2024scalable}, which requires training from scratch, and ShiftAddViT~\cite{you2024shiftaddvit}, which demands extensive parameter fine-tuning, ShiftAddLLM applies the shift-and-add concept to pre-trained LLMs without additional training or fine-tuning. We also use a multi-objective optimization and automated bit allocation strategy to further improve accuracy or reduce GPU latency, energy, and memory usage.

\section{Preliminaries}

\begin{wrapfigure}{r}{0.45\textwidth}
\begin{minipage}{0.45\textwidth}
\begin{algorithm}[H]
\caption{Alternating Multi-bit BCQ \cite{xu2018alternating}}
\label{alg:bcq}
\begin{algorithmic}[1]
\State \textbf{Input:} Full-precision weight $\mathbf{w} \in \mathbb{R}^n$, bit-width $q$, alternating cycles $T$
\State \textbf{Output:} $\alpha_i^*, \mathbf{b}_i^* \in \{-1, 1\}^{m\times n}$
\Function{Multi-bit BCQ}{$\mathbf{w}, q, T$}
    \State $\{\alpha_i, \mathbf{b}_i\}_{i=1}^{q} \gets \textsc{Greedy}(\mathbf{w})$ %
    \For{$t \gets 1 \text{ to } T$}
        \State $\{\alpha_i\}_{i=1}^{q} \gets \textsc{LS}(\mathbf{B}, \mathbf{w})$ %
        \State $\{\mathbf{b}_i\}_{i=1}^{q} \gets \textsc{BS}(\alpha_1, \ldots, \alpha_q, \mathbf{w})$
    \EndFor
\EndFunction
\end{algorithmic}
\end{algorithm}
\end{minipage}
\end{wrapfigure}

\textbf{Binary-coding Quantization (BCQ).} 
BCQ~\cite{xu2018alternating} quantizes each weight tensor
in an $L$-layer LLM
$\mathbf{w} \in \mathbb{R}^{m \times n}$ into $q$ bits using a linear combination of binary matrices $\{\mathbf{b}_i\}_{i=1}^{q}$ and corresponding scaling factors $\{\alpha_i\}_{i=1}^{q}$, where $\mathbf{b}_i \in \{-1, 1\}^{m \times n}$. The weights are then approximated by $\mathbf{w}_q = \sum_{i=1}^{q} \alpha_i \mathbf{b}_i$ as a result of minimizing the quantization error, i.e.,
$
    \arg \min_{\alpha_i, \mathbf{b}_i} \left\| \mathbf{w} - \sum_{i=1}^{q} \alpha_i \mathbf{b}_i \right\|^2
$
to obtain the optimal $\alpha_i^*, \mathbf{b}_i^*$.
If $q$ is 1, then the problem collapses to binary quantization, which has an analytical solution:
$
    \mathbf{b}^* = \text{sign}(\mathbf{w}), \alpha^* = \mathbf{w}^\top \mathbf{b}^* / n.
$
For multi-bit quantization, we resort to greedy and alternating methods~\cite{xu2018alternating,biq,kwon2020post}, as shown in Alg. \ref{alg:bcq}.
Initially, we use the greedy method~\cite{guo2017network} to initialize $\alpha_i, \mathbf{b}_i$, where the $i$-th bit quantization is performed by minimizing the residual $\mathbf{r}$ from the $(i-1)$-th bit:
\begin{equation}\label{eq:greedy}
\min_{\alpha_i, \mathbf{b}_i} \| \mathbf{r}_{i-1} - \alpha_i \mathbf{b}_i \|^2, \quad \text{where} \quad \mathbf{r}_{i-1} = \mathbf{w} - {\displaystyle\sum_{j=1}^{i-1}} \alpha_j \mathbf{b}_j, \quad 1 < i \leq q.
\end{equation}
We then obtain the initialized $\alpha_i, \mathbf{b}_i$ sequentially as $\mathbf{b}_i = \text{sign}(\mathbf{r}_i)$ and $\alpha_i = \mathbf{r}_i^\top \mathbf{b}_i / n$ \textbf{\textit{(Line 4)}}.
Next, we perform alternating optimization to further minimize the quantization error. Specifically, $\{\alpha_i\}_{i=1}^{q}$ can be iteratively refined using ordinary least squares (\textsc{LS})~\cite{guo2017network} as
$
    [\alpha_1, ..., \alpha_q] = ((\mathbf{B}^\top \mathbf{B})^{-1} \mathbf{B}^\top \mathbf{w})^\top,
$
where $\mathbf{B} = [\mathbf{b}_1, ..., \mathbf{b}_q] \in \{-1, 1\}^{m \times n \times q}$ \textbf{\textit{(Line 6)}}.
The binary codes $\{\mathbf{b}_i\}_{i=1}^{q}$ can then be iteratively recalibrated using a binary search (\textsc{BS}) given the refined $\{\alpha_i\}_{i=1}^{q}$ \textbf{\textit{(Line 7)}}~\cite{xu2018alternating}.

Such BCQ can support both uniform and non-uniform quantization formats by adjusting the scaling factors and biases accordingly~\cite{park2022lut}. Our ShiftAddLLM is built on top of BCQ but further replaces all associated multiplications with lower-cost hardware substitutes (e.g., shifts, adds, and LUT queries). We optimize not only the weight quantization error but also the output activation error, thereby achieving lower quantization bits along with savings in energy, memory, and computational costs.

\textbf{Shift and Add Primitives.}
Direct hardware implementation of multiplications is often inefficient. Using shift and add operations as ``shortcuts'' provides a more efficient alternative. Shifts, which are equivalent to multiplying by powers of two, offer a non-uniform quantization solution and can result in significant savings. For example, 
\hry{we tested matrix multiplication from one MLP layer of OPT-66B between weight $W \in \mathbb{R}^{9216 \times 36884}$ and activation $A \in \mathbb{R}^{1 \times 9216}$ using FP16 MACs and our 3-bit ShiftAddLLM. Energy consumption was 80.36J vs. 9.77J, achieving 87.8\% savings with our method.}
Both primitives have inspired many innovations in efficient model innovations~\cite{chen2020addernet,elhoushi2021deepshift,you2020shiftaddnet,you2024shiftaddvit}.

\section{The Proposed ShiftAddLLM Framework}

\textbf{Overview.}
We introduce our ShiftAddLLM as follows:
\textbf{\textit{First}}, we describe the reparameterization of pretrained LLMs through a \textit{post-training} shift-and-add approach in Sec. \ref{sec:shiftaddllm-1}. 
\textbf{\textit{Second}}, to enhance accuracy, we introduce a multi-objective optimization method that accounts for both weight quantization error and output activation error, detailed in Sec. \ref{sec:shiftaddllm-2}.
\textbf{\textit{Third}}, to improve efficiency, we explore a mixed and automated bit allocation strategy, illustrated in Sec. \ref{sec:shiftaddllm-3}.

\subsection{ShiftAddLLM: Post-training Reparameterization of LLMs with Shift and Add Primitives}
\label{sec:shiftaddllm-1}

\begin{figure}[t]
    \centering
    \includegraphics[width=0.99\linewidth]{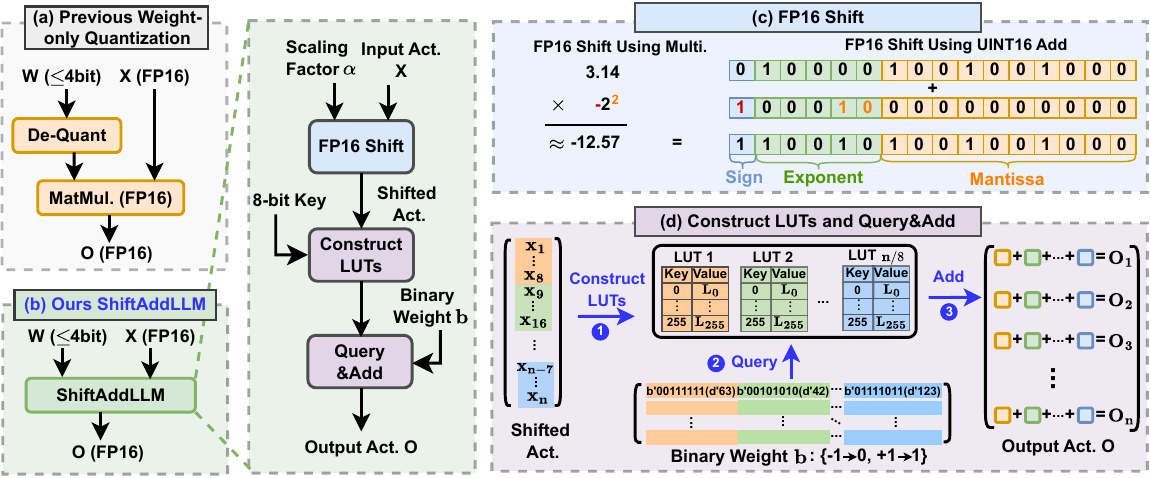}
    \caption{Illustration of our proposed post-training reparameterization for ShiftAddLLM.}
    \label{fig:shiftaddllm}
\end{figure}

\textbf{Post-training Reparameterization of LLMs.}
To avoid the need for fine-tuning after reparameterization, our method closely mimics the original multiplications used in LLMs. Previous methods, such as weight-only quantization techniques~\cite{frantar2022optq}, employ gradient-based or activation-aware uniform quantization to fit the pretrained weight distribution better, thereby achieving lower quantization errors. However, these methods often lack direct hardware support and require on-the-fly dequantization to \texttt{FP16} for multiplication with activations, as depicted in Fig. \ref{fig:shiftaddllm} (a). In contrast, our ShiftAddLLM uses the BCQ format, supporting non-uniform quantization with customized CUDA kernels~\cite{park2022lut,jeon2020biqgemm}, bypassing the need for dequantization, as illustrated in Fig. \ref{fig:shiftaddllm} (b). 
In particular, our method employs the Alg. \ref{alg:bcq} to quantize pretrained weights into binary matrices $\{\mathbf{b}_i\}_{i=1}^{q}$ and scaling factors $\{\alpha_i\}_{i=1}^{q}$. Note that during the alternating optimization cycles, we further quantize all scaling factors to powers of two (PoT)~\cite{li2019additive}, as described by the equation:
\begin{equation}\label{eq:apot}
\alpha_k = \textsc{POT}\left(\mathbf{r}_{k-1}\right) = \textsc{POT}(\alpha - \sum_{j=0}^{k-1} \alpha_j), \quad \text{where} \quad \textsc{POT}(\alpha) = \text{sign}(\alpha) \cdot 2^\mathbf{P}, \quad 1 \leq k \leq K.
\end{equation}
This additive PoT method adopts a greedy strategy to enhance the representational capacity of PoT, using $K$ scaling factors, where the $k$-th PoT minimizes the residual $\mathbf{r}$ of the $(k-1)$-th PoT. Each PoT effectively quantizes the scaling factor $\alpha$ into $\text{sign}(\alpha) \cdot 2^\mathbf{P}$, where $\text{sign}(\alpha)$ indicates sign flips, $\mathbf{P} = \text{round}(\log_2(\text{abs}(\alpha)))$, and $2^\mathbf{P}$ denotes a bitwise shift to the left ($\mathbf{P}>0$) or right ($\mathbf{P}<0$).

After the above reparameterization, we can then replace the associated multiplication between weights and activations into two steps:
(1) Bitwise shifts between activations and scaling factors. Note that the activation is still in the \texttt{FP16} format, and the multiplication between a floating-point number and a positive or negative PoT integer can be efficiently implemented by an integer addition instruction on existing hardware following DenseShift~\cite{li2023denseshift}, as also illustrated in Fig. \ref{fig:shiftaddllm} (c);
(2) Queries and adds intermediate shifted activations with the binary matrices. To implement this efficiently and reduce redundant additions or accumulations, as shown in Fig. \ref{fig:shiftaddllm} (d), we pre-compute 256 ($=2^8$) possible values for every eight elements in the shifted activations to construct LUTs. 
\hry{Here every eight grouped binary weights form an 8-bit key. Suppose the shifted activation is an $n$-dimensional vector. In that case, we will get $\nicefrac{n}{8}$ LUTs, where the grouped binary weights are used as keys, and the precomputed partial sums are stored as values.}
This allows us to handle the multiplication between the binary matrix $\mathbf{b}_i$ and the shifted activations as queries to the LUTs. We then add all the partial sums to obtain the final output activations \hry{in FP16 format}. Such LUTs are well supported by existing GPU kernels~\cite{park2022lut,jeon2020biqgemm}.
The reparameterization can be applied to all weights in pretrained LLMs in a \textit{post-training} manner, replacing costly multiplications with efficient hardware operations.

\textbf{\hry{Takeaway.}}
ShiftAddLLM presents a novel \textit{multiplication-less} approach that leverages non-uniform quantization via BCQ and additive PoT. This methodology enhances the representation capacity for outlier weights and activations of large magnitude compared to uniform quantization. Moreover, additive PoT effectively resolves the issue of limited quantization resolution for non-outlier weights and activations. Overall, it allows the quantization levels to better align with the data distribution.

\subsection{ShiftAddLLM: Multi-objective Optimization}
\label{sec:shiftaddllm-2}
\begin{figure}[t]
    \centering
    \includegraphics[width=\linewidth]{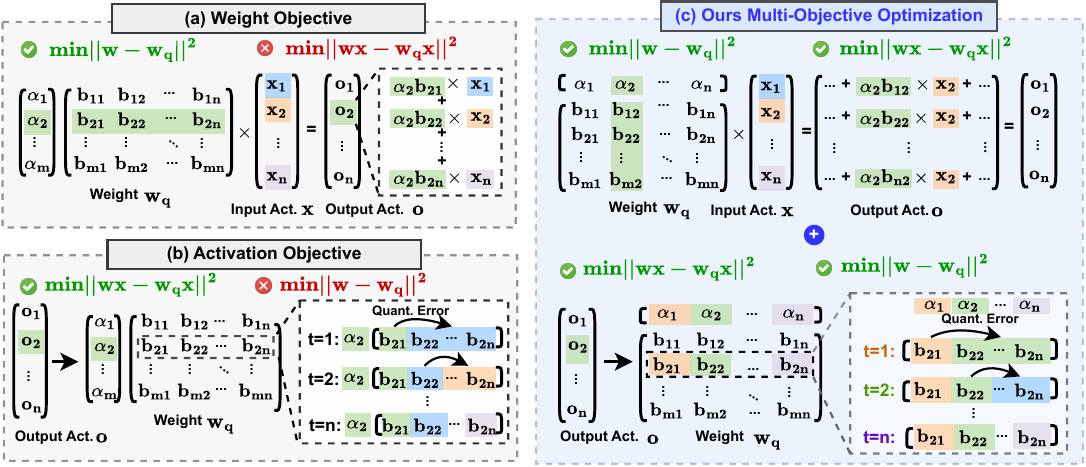}
    \caption{Illustration of our proposed multi-objective optimization framework.}
    \label{fig:multi_objective}
\end{figure}
\textbf{Motivating Analysis on Previous LLM Quantization Objectives.}
We examine previous weight-only quantization methods to understand the causes of large quantization error and accuracy drop.
These methods typically use either a \textit{weight} or \textit{activation} objective to minimize quantization error.
Specifically, the ``\textit{weight objective}'' (see Fig. \ref{fig:multi_objective} (a)) aims to minimize the weight quantization error, i.e.,
$
    \left\| \mathbf{W} - \mathbf{W}_q \right\|^2
$,
and adopts scaling factors for each row of quantized weights.
However, this does not optimize output activation error, as each weight element is multiplied by a unique input activation before summing to produce the output. 
Varying input activations, especially outliers~\cite{xiao2023smoothquant,lin2023awq}, rescale the weight quantization error differently, causing significant divergence in the output activation.
For example, LUT-GEMM~\cite{park2022lut} adopts this weight objective.
On the other hand, the ``\textit{activation objective}'' (see Fig. \ref{fig:multi_objective} (b)) minimizes the output activation error, i.e.,
$
    \left\| \mathbf{WX} - \mathbf{W}_q\mathbf{X} \right\|
$,
by quantizing one column of weights at a time and continuously updating the remaining unquantized weights to compensate for the quantization error incurred by quantizing a single weight column.
However, the fixed scaling factors may not adequately accommodate the weights adjusted afterward.
OPTQ~\cite{frantar2022optq} employs this activation objective.

\textbf{Our Multi-Objective Optimization.} 
To further mitigate accuracy drop after reparameterization (see Sec. \ref{sec:shiftaddllm-1}), we introduce a multi-objective optimization framework that combines weight and activation objectives using column-wise scaling factors. This framework effectively reduces quantization error for both weights and activations, thereby improving the accuracy of ShiftAddLLM.

As shown in Fig. \ref{fig:multi_objective} (c), using column-wise scaling factors overcomes the limitations of the previous weight objective~\cite{park2022lut} by eliminating the impact of varying input activations on quantized weights.
Each scaling factor corresponds to a constant activation value.
Additionally, scaling factors for subsequent columns are updated gradually after compensating for the corresponding column's weights, ensuring a better fit than the previous activation objective~\cite{frantar2022optq}.

\begin{wrapfigure}{r}{0.64\textwidth}
\setlength{\intextsep}{0pt} %
  \setlength{\columnsep}{0pt} %
  \captionsetup{aboveskip=0pt, belowskip=0pt}
  \centering
    \includegraphics[width=0.62\textwidth]{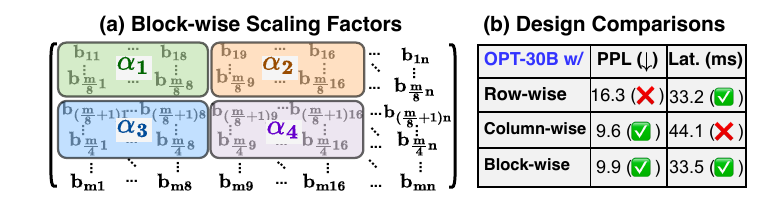}
  \caption{(a) the block-wise scaling factors and (b) the comparison among different designs on OPT-30B~\cite{zhang2022opt}.}
  \label{fig:block-wise}
\end{wrapfigure}%
\textbf{Accuracy vs. Latency Tradeoffs.}
The column-wise scaling factor design significantly boosts accuracy after reparameterization. However, it does not fully leverage BCQ~\cite{park2022lut,jeon2020biqgemm}, which process eight elements per row of weights in parallel as LUT keys, resulting in latency overhead for models with $\geq$30B parameters.
For example, testing on the OPT-30B~\cite{zhang2022opt} model and WikiText-2 dataset~\cite{merity2017pointer} showed \hry{(16.3 $-$ 9.6) $=$ 6.7} perplexity reduction but with a \hry{$\nicefrac{(44.1 - 33.2)}{44.1} \approx$ 24.7\%} latency overhead, as shown in Fig.~\ref{fig:block-wise} (b).

To address this, we propose a block-wise scaling factor design that groups 8 columns and \nicefrac{1}{8} of the original rows to share a scaling factor, ensuring compatibility with the BCQ kernel and achieving latency reductions, as shown in Fig. \ref{fig:block-wise} (a). 
We refer to ShiftAddLLM with column-wise scaling factors as ``\textit{Ours (Acc.)}'' for high accuracy optimization, and with block-wise scaling factors as ``\textit{Ours (Lat.)}'' for optimized accuracy-latency trade-off.

\textbf{\hry{Takeaway.}}
Our multi-objective optimization approach integrates both weight and activation objectives, reducing weight quantization error in an activation-aware manner and output activation error reduction in a weight-aware manner. This synergy, achieved through a simple column-wise or block-wise design, significantly boosts the accuracy of weight-only quantization. This aligns with the principles of previous activation-aware weight quantization methods~\cite{lin2023awq}.

\subsection{ShiftAddLLM: Mixed and Automated Bit Allocation}
\label{sec:shiftaddllm-3}

\begin{wrapfigure}{r}{0.56\textwidth}
  \centering
    \includegraphics[width=0.55\textwidth]{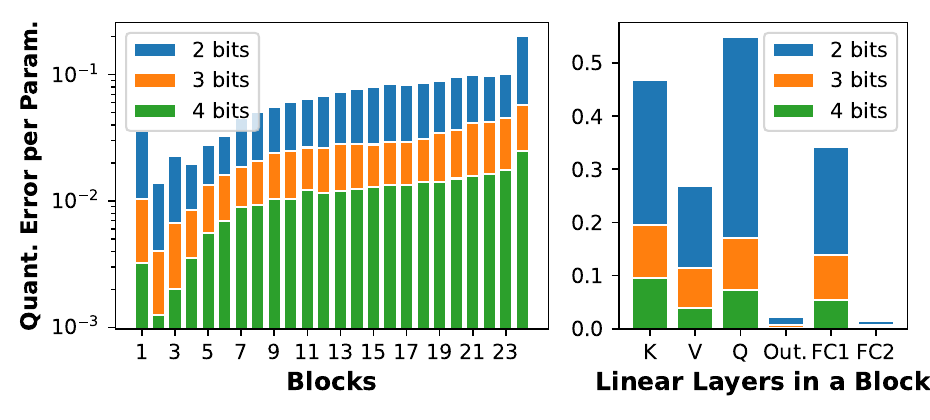}
  \caption{\hry{Sensitivity analysis on OPT-1.3B~\cite{zhang2022opt}.}
  }
  \label{fig:sensitivity}
\end{wrapfigure}%

\textbf{Sensitivity Analysis.}
We analyze the sensitivity of different layers and blocks in LLMs to shift-and-add reparameterization. As shown in Fig. \ref{fig:sensitivity}, later blocks incur more quantization or reparameterization errors. Within each block, \hry{Query/Key (Q/K) layers} are generally more sensitive to reparameterization than other linear layers.
This diverse sensitivity motivates us to explore mixed bit allocations for LLM reparameterization and develop strategies to automatically determine the optimal bit allocations given the average bit budgets.

\begin{wrapfigure}{r}{0.3\textwidth}
  \centering
    \includegraphics[width=0.3\textwidth]{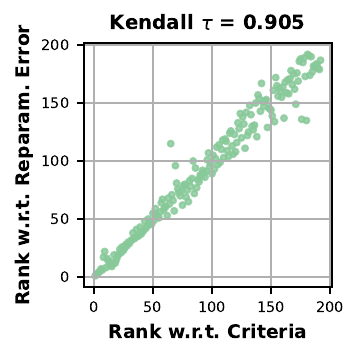}
  \caption{Rank comparisons.}
  \label{fig:ranking}
\end{wrapfigure}%
\textbf{Criteria and Automated Bit Allocation.}
To develop the bit allocation scheme, we propose criteria to estimate the importance of linear weights and formulate the bit allocation as an integer programming problem. For weight $\mathbf{W}_i$ from the $i$-th layer of an LLM, the criterion $C_i$ is defined as follows:
\begin{equation}\label{eq:criteria}
\begin{split}
C_i & = \| \text{IS} \|_F \cdot \text{STD}(\text{IS})^2, \quad \text{where} \quad \\
\text{IS} & = \mathbf{W}_i / \text{diag}(\text{cholesky}((\mathbf{X}_i\mathbf{X}_i^T)^{-1})), 
\end{split}
\end{equation}
where the importance score (IS) is inspired by Optimal Brain Compression~\cite{hassibi1993optimal,frantar2022optimal,frantar2022optq} and is correlated to the increase in the quadratic reconstruction error $\left\| \mathbf{WX} - \mathbf{W}_q\mathbf{X} \right\|^2$ after reparameterizing the weights, i.e., IS $\uparrow$, error increases $\downarrow$. The $F$-norm of IS indicates the overall importance of \( \mathbf{W}_i \), while the standard deviation (STD) highlights the reparameterization difficulty for outliers. Considering both factors, we achieve a more effective evaluation metric proportional to the actual reparameterization error. As shown in Fig. \ref{fig:ranking}, the rankings derived from our defined criteria and the actual reparameterization error are highly correlated, with a Kendall $\tau$ of 0.905.
To refine the criteria by incorporating the bit-width, we use least squares polynomial fits to estimate each bit's corresponding reparameterization error as $C_{i,b}$.

Given the criteria, we can formulate the automated bit allocation as an integer programming problem:
\begin{equation} \label{eq:mix_bit}
\arg \min_{\beta_{i,b}} \sum_{i}^L \sum_{b} \beta_{i,b} \cdot C_{i,b}, \quad
\text{s.t.} \quad \sum_{b} \beta_{i,b} = 1, \quad
\sum_{i}^L \sum_{b} \beta_{i,b} \cdot b \leq \mathcal{B} \cdot L, 
\end{equation}
where $L$ is the number of layers in the target LLM, $b \in \{2,3,4\}$ denotes the available bit widths, and $\beta_{i,b} \in \{0,1\}$ is the one-hot indicator for the $i$-th layer to determine the assigned bits, e.g., $\{0, 1, 0\}$ means 3 bits. The objective is to minimize the summed criteria $C$ of all layers under the given average bit budget $\mathcal{B}$ per layer. The final $\beta_{i,b}$ represents the assigned bits for the $i$-th layer in the target LLM.

\textbf{\hry{Takeaway.}}
Using mixed bits instead of static ones can improve the accuracy-efficiency tradeoffs by adapting the varying sensitivities across layers, e.g., Q/K linear layers exhibit higher sensitivity to reparameterization; Our adopted criteria provide a quick estimation of the reparameterization error.

\section{Experiments}
\label{sec:exps}

\subsection{Experiment Settings}
\label{sec:exps_setting}

\textbf{\textit{Models.}}
We consider five representative SOTA LLM families, including OPT~\cite{zhang2022opt}, LLaMA-1/2/3~\cite{touvron2023llama2,llama3}, Gemma~\cite{team2024gemma}, Mistral~\cite{jiang2023mistral}, and Bloom~\cite{le2023bloom}.
\textbf{\textit{Tasks and Datasets.}} We evaluate all five LLMs on the commonly adopted language modeling task using the WikiText-2~\cite{merity2017pointer} dataset for perplexity measurement. Additionally, we extend the evaluation of the two largest models, OPT-66B and LLaMA-2-70B, to \hry{eight} downstream tasks for zero-shot accuracy evaluation. These tasks include ARC (Challenge/Easy)~\cite{boratko2018systematic}, BoolQ~\cite{clark2019boolq}, 
Copa~\cite{afshar2018copa}, PIQA~\cite{tata2003piqa}, RTE~\cite{dagan2022recognizing},  StoryCloze~\cite{mostafazadeh2017lsdsem}, \hry{and MMLU~\cite{hendrycks2020measuring}.}
\textbf{\textit{Baselines.}} We consider four SOTA LLM quantization methods: OPTQ~\cite{frantar2022optq}, LUT-GEMM~\cite{park2022lut}, QuIP~\cite{chee2024quip}, and AWQ~\cite{lin2023awq}.
\textbf{\textit{Evaluation Metrics.}} We evaluate ShiftAddLLM and the baselines using both accuracy and efficiency metrics. For accuracy, we evaluate perplexity on the WikiText-2 dataset and zero-shot accuracy on \hry{eight} downstream tasks. For efficiency, we measure the latency on a single A100-80GB GPU (PCIe)~\cite{nvidia2020a100} and estimate the energy costs using an Eyeriss-like hardware accelerator~\cite{chen2016eyeriss,zhao2020dnn}, which calculates not only computational but also data movement energy \hry{(within 18\% of the differences with Eyeriss’s chip measurement results as claimed)}.

\begin{table}[t]
\centering
\renewcommand{\arraystretch}{1.2}
\caption{Perplexity comparisons of the OPT models on WikiText-2. Note that we set the group size of all methods as the length of rows following the setting of OPTQ~\cite{frantar2022optq} for a fair comparison.
}
\resizebox{\linewidth}{!}{
\begin{tabular}{l|c|cccccccccc}
\Xhline{3\arrayrulewidth}
\textbf{OPT (PPL $\downarrow$)} & \textbf{Bits} & \textbf{125M} & \textbf{350M} & \textbf{1.3B} & \textbf{2.7B} & \textbf{6.7B} & \textbf{13B} & \textbf{30B} & \textbf{66B} \\
\Xhline{3\arrayrulewidth}
FP16 & 16 & 27.65 & 22.00 & 14.62 & 12.47 & 10.86 & 10.13 & 9.56 & 9.34 \\
\hline
OPTQ~\cite{frantar2022optq} & 3 & 53.85 & 33.79 & 20.97 & 16.88 & 14.86 & 11.61 & 10.27 & 14.16 \\
LUT-GEMM~\cite{park2022lut} & 3 & 60.00 & 42.32 & 49.10 & 17.55 & 17.44 & 12.50 & 139.90 & 100.33 \\
AWQ~\cite{lin2023awq} & 3 & 54.75 & 35416.00 & 24.60 & 39.01 & 16.47 & 16.53 & 31.01 & 5622.00 \\
\textbf{Ours (Acc.)} & 3 & \textbf{31.29} & \textbf{24.24} & \textbf{21.53} & \textbf{13.68} & \textbf{11.18} & \textbf{10.39} & \textbf{9.63} & \textbf{9.43} \\
\Xhline{3\arrayrulewidth}
OPTQ~\cite{frantar2022optq} & 2 & 2467.50 & 10433.30 & 4737.05 & 6294.68 & 442.63 & 126.09 & 71.70 & 20.91 \\
LUT-GEMM~\cite{park2022lut} & 2 & 4844.32 & 2042.90 & 3851.50 & 616.30 & 17455.52 & 4963.27 & 7727.27 & 6246.00 \\
AWQ~\cite{lin2023awq} & 2 & 3514.61 & 18313.24 & 9472.81 & 22857.70 & 8168.30 & 5014.92 & 7780.96 & 103843.84 \\
QuIP~\cite{chee2024quip} & 2 & 92.84 & 146.15 & 27.90 & 30.02 & 16.30 & 12.34 & 11.48 & 10.92 \\
\textbf{Ours (Acc.)} & 2 & \textbf{51.15} & \textbf{40.24} & \textbf{29.03} & \textbf{20.78} & \textbf{13.78} & \textbf{12.17} & \textbf{10.67} & \textbf{10.33} \\
\Xhline{3\arrayrulewidth}
\end{tabular}
}
\label{tab:opt_ppl}
\end{table}

\subsection{ShiftAddLLM over SOTA LLM Quantization Baselines}

\begin{figure}[t]
    \centering
    \includegraphics[width=\linewidth]{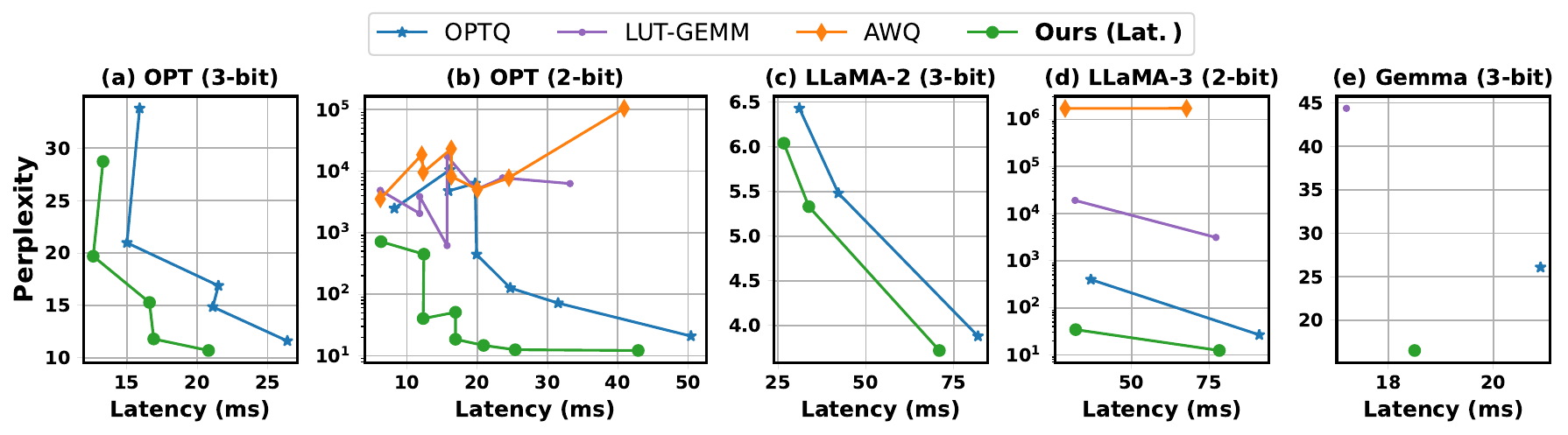}
    \caption{Accuracy-latency tradeoff comparisons on the OPT, LLaMA-2/3, and Gemma models.}
    \label{fig:ours_lat}
\end{figure}

\textbf{Results on OPT Models.}
To evaluate the effectiveness of our ShiftAddLLM, we compare against four SOTA LLM quantization baselines: OPTQ~\cite{frantar2022optq}, LUT-GEMM~\cite{park2022lut}, QuIP~\cite{chee2024quip}, and AWQ~\cite{lin2023awq}. 
Using the OPT model family~\cite{zhang2022opt} and the WikiText-2 dataset~\cite{merity2017pointer}, we assess perplexity, GPU latency, and energy costs.
As shown in Tab.~\ref{tab:opt_ppl}, \textit{Ours (Acc.)} consistently outperforms all baselines, achieving an average perplexity reduction of \hr{5.63/38.47/5136.13} compared to OPTQ, LUT-GEMM, and AWQ, respectively, at 3 bits. 
At 2 bits, where most baselines fail with significantly high perplexity, our method maintains low perplexity, and achieves an average \hr{22.74} perplexity reduction over the most competitive QuIP.
Also, as shown in Fig. \ref{fig:ours_lat} (a \& b), \textit{Ours (Lat.)} consistently achieves better accuracy-latency tradeoffs, with a perplexity reduction of \hr{0.91$\sim$103830.45} at comparable latency or \hr{6.5\%$\sim$60.1\%} latency reductions and \hr{26.0\%$\sim$44.7\%} energy savings at similar or even lower perplexity.
Complete quantitative data on accuracy, latency, and energy is provided in Appendix~\ref{sec:opt_appendix}.

\begin{wraptable}{r}{0.63\textwidth}
\centering
\setlength{\tabcolsep}{3.5pt}
\renewcommand{\arraystretch}{1.2}
\caption{Perplexity comparisons of the LLaMA models on WikiText-2. The group size is set to 128 following \cite{park2022lut,lin2023awq}.}
\resizebox{\linewidth}{!}{
\begin{tabular}{l|c|c|x{1cm}x{1cm}x{0.9cm}|x{0.9cm}x{0.9cm}}
\Xhline{3\arrayrulewidth}
\multirow{2}[1]{*}{\textbf{LLaMA (PPL $\downarrow$)}} & \multirow{2}[1]{*}{\textbf{Bits}} & \multicolumn{1}{c|}{\textbf{LLaMA-1}} & \multicolumn{3}{c|}{\textbf{LLaMA-2}} & \multicolumn{2}{c}{\textbf{LLaMA-3}} \\
\cline{3-8}      &   & \textbf{7B} & \textbf{7B} & \textbf{13B} & \textbf{70B} & \textbf{8B} & \textbf{70B} \\
\Xhline{3\arrayrulewidth}
FP16 & 16 & 5.68 & 5.47 & 4.88 & 3.32 & 6.14 & 2.86 \\
\hline
OPTQ~\cite{frantar2022optq} & 3 & 8.81 & 6.43 & 5.48 & 3.88 & 13.69 & 4.91 \\
LUT-GEMM~\cite{park2022lut} & 3 & 7.18 & 7.02 & 5.89 & 4.01 & 11.10 & 5.92 \\
AWQ~\cite{lin2023awq} & 3 & 6.35 & 6.24 & 5.32 & 3.74 & 8.15 & 4.69 \\
\textbf{Ours (Acc.)} & 3 & \textbf{6.04} & \textbf{5.89} & \textbf{5.16} & \textbf{3.64} & \textbf{7.20} & \textbf{4.35}\\
\Xhline{3\arrayrulewidth}
OPTQ~\cite{frantar2022optq} & 2 & 68.60 &  19.92 & 12.75 & 6.82 & 398.0 & 26.65 \\
LUT-GEMM~\cite{park2022lut} & 2 & 303.00 & 2242.0 & 2791.0 & 136.4 & 19096 & 3121 \\
AWQ~\cite{lin2023awq} & 2 & 2.6e5 & 2.2e5 & 1.2e5 & 7.2e4 & 1.7e6 & 1.7e6 \\
\textbf{Ours (Acc.)} & 2 & \textbf{7.98} & \textbf{8.51} & \textbf{6.77} & \textbf{4.72} & \textbf{12.07} & \textbf{7.51} \\
\Xhline{3\arrayrulewidth}
\end{tabular}
}
\label{tab:LLaMA_ppl}
\end{wraptable}

\textbf{Results on LLaMA Models.}
We further evaluate ShiftAddLLM on LLaMA models~\cite{touvron2023llama,touvron2023llama2,llama3} due to their superior performance among open-source LLMs.
As shown in Tab. \ref{tab:LLaMA_ppl}, \textit{Ours (Acc.)} consistently outperforms all baselines, achieving an average perplexity reduction of \hr{1.82/1.47/0.29} and \hr{80.87/4606.98/678658.74} compared to OPTQ, LUT-GEMM, and AWQ at 3 and 2 bits, respectively.
Evaluating \textit{Ours (Lat.)} with both accuracy and latency metrics as shown in Fig. \ref{fig:ours_lat} (c \& d), \textit{Ours (Lat.)} demonstrates better accuracy-latency tradeoffs.
It achieves \hr{1.1$\sim$1719987.6} perplexity reduction at comparable latency or \hr{19.9\%$\sim$65.0\%} latency reductions and \hr{28.4\%$\sim$89.9\%} energy savings at similar or even lower perplexity.
Complete quantitative data on accuracy, latency, and energy are provided in Appendix~\ref{sec:llama_appendix}.

\begin{wraptable}{r}{0.61\textwidth}
\centering
\setlength{\tabcolsep}{3.6pt}
\renewcommand{\arraystretch}{1.2}
\caption{Results on Gemma/Mistral/Bloom models.}
\resizebox{\linewidth}{!}{
\begin{tabular}{l|c|cccc}
\Xhline{3\arrayrulewidth}
\textbf{PPL ($\downarrow$)} & \textbf{Bits} & \textbf{Gemma-2B} & \textbf{Mistral-7B} & \textbf{Bloom-3B} & \textbf{Bloom-7B} \\
\Xhline{3\arrayrulewidth}
FP16 & 16 & 13.88 & 5.25 & 13.48 & 11.37 \\
\hline
OPTQ & 3 & 26.08 & 7.27 & 17.40 & 13.47 \\
LUT-GEMM & 3 & 44.36 & 22.36 & 21.03 & 17.29 \\
\textbf{Ours (Acc.)} & 3 & \textbf{14.96} & \textbf{5.60} & \textbf{14.10} & \textbf{11.71} \\
\Xhline{3\arrayrulewidth}
\end{tabular}
}
\label{tab:other_ppl}
\end{wraptable}

\textbf{Results on Gemma/Mistral/Bloom Models.}
We also evaluate ShiftAddLLM on Gemma~\cite{team2024gemma}, Mistral~\cite{jiang2023mistral}, and Bloom~\cite{le2023bloom} models, which are among the most popular open-source LLMs and Mixture-of-Expert (MoE) models.
As shown in Tab.~\ref{tab:other_ppl}, \textit{Ours (Acc.)} achieves perplexity reductions of \hr{11.12/29.4} for Gemma-2B, \hr{1.67/16.76} for Mistral-7B, and \hr{3.30/6.93} and \hr{1.76/5.58} for BLOOM-3B/7B, respectively, compared to OPTQ and LUT-GEMM.
As shown in Fig. \ref{fig:ours_lat} (e), \textit{Ours (Lat.)} shows better accuracy-latency tradeoffs, e.g., achieving \hr{9.56} perplexity reduction and \hr{11\%} latency reductions over the OTPQ baseline on Gemma models.
These results on five LLM families consistently validate the effectiveness of our ShiftAddLLM.

\textbf{Zero-shot Downstream Tasks.}
We extend our evaluation to zero-shot downstream datasets for a more comprehensive assessment.
As shown in Tab. \ref{tab:downstream_task}, \textit{Ours (Acc.)} consistently improves performance over previous SOTA baselines, achieving an average accuracy gain of \hry{13.37/13.19 and 2.55/2.39} over OPTQ and LUT-GEMM baselines at 3 bits when evaluated on OPT-66B and LLaMA-2-70B, respectively.
These experiments demonstrate that our method not only reduces perplexity but also improves downstream task accuracy.

\begin{table}[t]
\centering
\setlength{\tabcolsep}{4pt}
\renewcommand{\arraystretch}{1.2}
\caption{\hry{Accuracy comparisons on seven downstream tasks for OPT-66B and LLaMA-2-70B.}}
\resizebox{\linewidth}{!}{
\begin{tabular}{l|l|c|cccccccc|c}
\Xhline{3\arrayrulewidth}
\textbf{Models} & \textbf{Methods} & \textbf{Bits} & \textbf{ARC\_C} & \textbf{ARC\_E} & \textbf{Copa} & \textbf{BoolQ} & \textbf{PIQA} & \textbf{Storycloze} & \textbf{RTE} & \hry{\textbf{MMLU}} & \textbf{Mean} \\
\Xhline{3\arrayrulewidth}
\multirow{3}{*}{\tabincell{l}{OPT-66B}} 
& \hry{Floating Point} & 16 & 37.20 & 71.25 & 86 & 69.82 & 78.67 & 77.47 & 60.65 & 25.89$\pm$0.37 & 63.37 \\
& OPTQ~\cite{frantar2022optq} & 3 & 24.66 & 48.86 & 70 & 52.05 & 64.47 & 67.09 & 53.07 & 23.98$\pm$0.36 & 50.52 \\
& LUT-GEMM~\cite{park2022lut} & 3 & 24.15 & 51.85 & 81 & 53.52 & 61.97 & 60.60 & 48.74 & 23.73$\pm$0.36 & 50.70 \\
& \textbf{Ours (Acc.)} & 3 & \textbf{35.24} & \textbf{70.88} & \textbf{87} & \textbf{72.45} & \textbf{77.64} & \textbf{77.15} & \textbf{63.18} & \textbf{27.56$\pm$0.38} & \textbf{63.89}  \\
\hline
\multirow{3}{*}{\tabincell{l}{LLaMA-2-70B}} 
& \hry{Floating Point} & 16 & 49.57 & 76.14 & 90 & 82.57 & 80.79 & 78.61 & 68.23 & 65.24$\pm$0.37 & 72.89 \\
& OPTQ~\cite{frantar2022optq} & 3 & 45.82 & 76.34 & 90 & 81.74 & 79.71 & 77.34 & 67.51 & 60.14$\pm$0.36 & 72.33 \\
& LUT-GEMM~\cite{park2022lut} & 3 & 47.70 & 76.42 & 89 & 80.31 & 80.20 & 77.78 & 68.59 & - & - \\
& \textbf{Ours (Acc.)} & 3 & \textbf{48.38} & \textbf{77.06} & \textbf{93} & \textbf{84.25} & \textbf{80.47} & \textbf{78.49} & \textbf{75.09} & \textbf{62.33$\pm$0.38} & \textbf{74.88} \\
\Xhline{3\arrayrulewidth}
\end{tabular}
}
\label{tab:downstream_task}
\end{table}

\textbf{GPU Memory Savings.}
Our ShiftAddLLM also reduces GPU memory usage. 
For OPT-66B, our method saves \hr{81\% and 87\%} memory costs over \texttt{FP16} at 3 (23GB vs. 122GB) and 2 bits (16GB vs. 122GB), respectively.
For LLaMA-2-70B, it saves \hr{80\% and 87\%} memory costs at 3 (25GB vs. 128GB) and 2 bits (17GB vs. 128GB), respectively.
\begin{table}[t]
    \centering
    \setlength{\tabcolsep}{4pt}
    \renewcommand{\arraystretch}{1.2}
    \caption{Perplexity and latency results of our mixed bit allocation.}
    \resizebox{\linewidth}{!}{
    \begin{tabular}{l|c|cccccc|cccccc}
        \Xhline{2\arrayrulewidth}
        \multirow{2}[1]{*}{\textbf{Methods}} & \multirow{2}[1]{*}{\textbf{Bits}} & \multicolumn{6}{c|}{\textbf{PPL ($\downarrow$)}} & \multicolumn{6}{c}{\textbf{Latency (ms)}} \\
        \cline{3-14}
        & & \textbf{125M} & \textbf{350M} & \textbf{1.3B} & \textbf{2.7B} & \textbf{6.7B} & \textbf{13B} & \textbf{125M} & \textbf{350M} & \textbf{1.3B} & \textbf{2.7B} & \textbf{6.7B} & \textbf{13B} \\
        \Xhline{2\arrayrulewidth}
        \textbf{Ours (Lat.)} & 2 & 712.55 & 445.78 & 40.28 & 50.95 & 18.56 & 14.76 & 6.3 & 12.4 & 12.3 & 16.9 & 16.9 & 20.9 \\
        \textbf{Ours (Mixed)} & 2.2 & 435.84 & 279.19 & 27.37 & 31.97 & 17.99 & 13.79 & 6.3 & 12.6 & 12.5 & 16.8 & 16.7 & 21.0 \\
        \Xhline{2\arrayrulewidth}
    \end{tabular}
    }
    \label{tab:mixed_bit}
\end{table}

\textbf{Results of Mixed Bit Allocation.}
We evaluate our mixed bit allocation strategy (see Sec. \ref{sec:shiftaddllm-3}) and compare \textit{Ours (Mixed)} with \textit{Ours (Lat.)}.
As shown in Tab.~\ref{tab:mixed_bit}, \textit{Ours (Mixed)} further reduces the perplexity by an average of \hr{79.45} for OPT model families under comparable or even less latency.
\hry{We provide more results in Appendix~\ref{sec:more_mixed_bit} to validate the effectiveness of our mixed bit allocation strategy.}

\subsection{Ablation Studies of ShiftAddLLM}
\begin{wrapfigure}{r}{0.5\textwidth}
  \centering
    \includegraphics[width=0.5\textwidth]{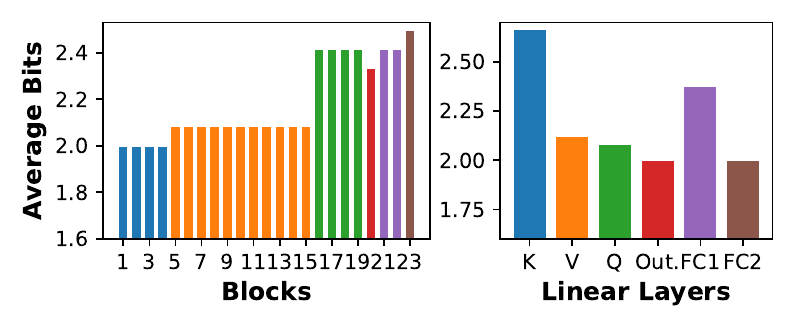}
  \caption{\hry{Visualizing the average bit allocation.}}
  \label{fig:bit_avg}
\end{wrapfigure}%

\textbf{Visualization of Mixed Bit Allocation.}
We visualize the bit allocations after applying our automated bit allocation strategy with an average bit budget of 2.2 (Fig. \ref{fig:bit_avg}). 
The allocation pattern correlates with the sensitivity to reparameterization identified in Sec. \ref{sec:shiftaddllm-3} and shown in Fig. \ref{fig:sensitivity}.
For instance, later blocks, which experience more quantization or reparameterization errors, receive more bits. The K linear layers and the first MLP (FC1) in each block are also allocated higher bits. This visualization confirms that our strategy effectively adjusts bits according to reparameterization errors.

\begin{wraptable}{r}{0.45\textwidth}
\centering
\setlength{\tabcolsep}{3.6pt}
\renewcommand{\arraystretch}{1.2}
\caption{Performance breakdown analysis.}
\resizebox{\linewidth}{!}{
\begin{tabular}{l|c|cc|cc}
\Xhline{3\arrayrulewidth}
\multirow{2}[1]{*}{\textbf{OPT w/ Sec.}} & \multirow{2}[1]{*}{\textbf{Bits}} & \multicolumn{2}{c|}{\textbf{PPL}} & \multicolumn{2}{c}{\textbf{Latency (ms)}} \\
\cline{3-6}
& \textbf{} & \textbf{6.7B} & \textbf{13B} & \textbf{6.7B} & \textbf{13B} \\
\Xhline{3\arrayrulewidth}
\ref{sec:shiftaddllm-1} & 2 & 6.4e4 & 1.5e4 & 16.5 & 20.1 \\
\ref{sec:shiftaddllm-1}\&\ref{sec:shiftaddllm-2} & 2 & 18.56 & 14.76 & 16.9 & 20.9 \\
\ref{sec:shiftaddllm-1}\&\ref{sec:shiftaddllm-2}\&~\ref{sec:shiftaddllm-3} & 2.2 & \textbf{17.99} & \textbf{13.79} & 16.7 & 21.0 \\
\Xhline{3\arrayrulewidth}
\end{tabular}
}
\label{tab:opt_breakdown}
\end{wraptable}

\begin{figure}[t]
    \centering
    \includegraphics[width=\linewidth]{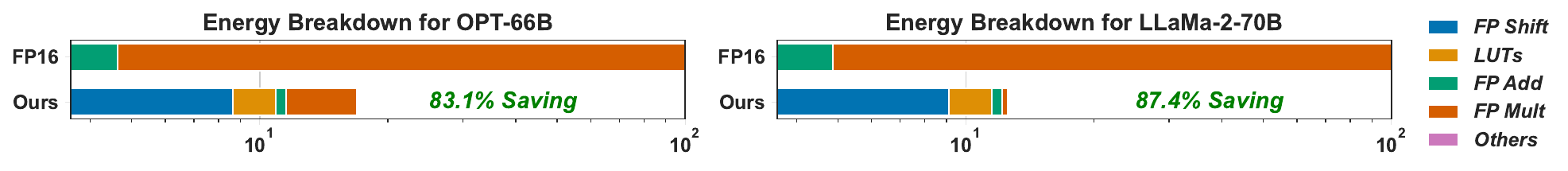}
    \caption{\hry{Energy breakdown for OPT-66B and LLaMA-70B models using an Eyeriss accelerator.}}
    \label{fig:energy}
\end{figure}

\textbf{Performance and Energy Breakdown.}
To examine the contribution of each proposed technique, we conducted ablation studies on OPT-6.7B/13B models.
As shown in Tab. \ref{tab:opt_breakdown}, the vanilla ShiftAddLLM (Sec.~\ref{sec:shiftaddllm-1}) suffers from a significant perplexity increase with 2-bit reparameterization.
Our multi-objective optimization (Sec.~\ref{sec:shiftaddllm-2}) reduces perplexity by an average of \hr{3.9e4}, and the mixed bit allocation strategy (Sec.~\ref{sec:shiftaddllm-3}) further reduces perplexity by \hr{0.77}, maintaining comparable latency. 
These experiments validate the effectiveness of each component in ShiftAddLLM.
In addition, profiling the two largest models on an Eyeriss accelerator illustrates the energy breakdown of the original LLMs and ShiftAddLLMs.
As shown in Fig. \ref{fig:energy}, ShiftAddLLM reduces energy consumption by \hr{87.2\%} for OPT-66B and \hr{86.0\%} for LLaMa-2-70B, with shift-and-add leading to \hr{89.7\%} and \hr{89.9\%} energy reduction compared to original multiplications. 

\subsection{\hry{Discussion on Limitation}}
\label{sec:limitation}

We demonstrated the accuracy and efficiency of post-training shift-and-add reparameterization of LLMs using multi-objective optimization and automated bit allocation, addressing the challenge of efficient LLM serving. However, achieving GPU speedup relied on BCQ kernels and the compatible Ours (Lat.) with a block-wise scaling factor design. While Ours (Acc.) with a column-wise design delivers high accuracy, we lack the fast CUDA kernel required for similar speedups.

\subsection{\hry{Discussion on Technique Applicability Beyond LLMs}}
\label{sec:discussion}

We acknowledge that the idea of shift-and-add reparameterization is general and can be extended to other smaller models like CNNs~\cite{you2020shiftaddnet} or ViTs~\cite{you2024shiftaddvit}. Meanwhile, this work’s implementation is specifically dedicated to large-scale LLMs: It is the first instance of applying the shift-and-add technique at the scale of LLMs with billions of parameters. While many ideas perform well with models having millions of parameters, they often fail to scale effectively. Unlike previous methods that require additional training and do not yield good results for large-scale LLMs, our approach is uniquely tailored for LLMs. We incorporate ``post-training'' reparameterization and carefully designed scaling factor patterns, enabling multi-objective optimization for LLMs and ensuring superior performance compared to prior quantization methods.

\section{Conclusion}

We propose accelerating pretrained LLMs through post-training shift-and-add reparameterization, creating efficient multiplication-free models. Our method reparameterizes weight matrices into binary matrices with group-wise scaling factors, transforming multiplications into shifts and adds.
To mitigate accuracy loss, we introduce a multi-objective optimization strategy that minimizes weight and activation reparameterization errors. 
Additionally, we develop an automated bit allocation strategy based on layer sensitivity to further improve the accuracy-efficiency tradeoff. Extensive results across various LLM families and tasks validate the effectiveness of ShiftAddLLM. This work opens a new perspective on designing efficient LLM serving systems through \textit{post-training} optimization.

\begin{ack}
This work is supported by the National Science Foundation (NSF) RTML program (Award number: 1937592) and the CoCoSys, one of the seven centers in JUMP 2.0, a Semiconductor Research Corporation (SRC) program sponsored by DARPA.
We extend our gratitude towards Mitchelle Rasquinha, and Robert Hundt for reviewing the paper and providing insightful feedback. We also thank the extended team at Google DeepMind who enabled and supported this research direction. 
\end{ack}

{
\small
\bibliographystyle{plainnat}
\bibliography{ref}

\begin{thebibliography}{76}
\providecommand{\natexlab}[1]{#1}
\providecommand{\url}[1]{\texttt{#1}}
\expandafter\ifx\csname urlstyle\endcsname\relax
  \providecommand{\doi}[1]{doi: #1}\else
  \providecommand{\doi}{doi: \begingroup \urlstyle{rm}\Url}\fi

\bibitem[Afshar et~al.(2018)Afshar, Perros, Papalexakis, et~al.]{afshar2018copa}
Ardavan Afshar, Ioakeim Perros, Evangelos~E Papalexakis, et~al.
\newblock {COPA: Constrained PARAFAC2 for sparse \& large datasets}.
\newblock In \emph{CIKM}, 2018.

\bibitem[AI(2024)]{llama3}
Meta AI.
\newblock {LLaMA 3}.
\newblock \url{https://github.com/meta-llama/llama3}, 2024.

\bibitem[Anil et~al.(2023)Anil, Borgeaud, et~al.]{team2023gemini}
Rohan Anil, Sebastian Borgeaud, et~al.
\newblock {Gemini: A Family of Highly Capable Multimodal Models}.
\newblock \emph{arXiv preprint arXiv:2312.11805}, 2023.

\bibitem[Boratko et~al.(2018)Boratko, Padigela, Mikkilineni, et~al.]{boratko2018systematic}
Michael Boratko, Harshit Padigela, Divyendra Mikkilineni, et~al.
\newblock {A Systematic Classification of Knowledge, Reasoning, and Context within the ARC Dataset}.
\newblock \emph{arXiv preprint arXiv:1806.00358}, 2018.

\bibitem[Brito et~al.(2014)Brito, Rabuske, Fernandes, et~al.]{brito2014quaternary}
Diogo Brito, Taimur~G Rabuske, Jorge~R Fernandes, et~al.
\newblock {Quaternary logic lookup table in standard CMOS}.
\newblock \emph{TVLSI}, 2014.

\bibitem[Chee et~al.(2024)Chee, Cai, Kuleshov, et~al.]{chee2024quip}
Jerry Chee, Yaohui Cai, Volodymyr Kuleshov, et~al.
\newblock {QuIP: 2-Bit Quantization of Large Language Models With Guarantees}.
\newblock \emph{NeurIPS}, 2024.

\bibitem[Chen et~al.(2020)Chen, Wang, Xu, et~al.]{chen2020addernet}
Hanting Chen, Yunhe Wang, Chunjing Xu, et~al.
\newblock {AdderNet: Do We Really Need Multiplications in Deep Learning?}
\newblock In \emph{CVPR}, 2020.

\bibitem[Chen et~al.(2016)Chen, Krishna, Emer, et~al.]{chen2016eyeriss}
Yu-Hsin Chen, Tushar Krishna, Joel~S Emer, et~al.
\newblock {Eyeriss: An Energy-efficient Reconfigurable Accelerator for Deep Convolutional Neural Networks}.
\newblock \emph{JSSCC}, 2016.

\bibitem[Clark et~al.(2019)Clark, Lee, Chang, et~al.]{clark2019boolq}
Christopher Clark, Kenton Lee, Ming-Wei Chang, et~al.
\newblock {BoolQ: Exploring the Surprising Difficulty of Natural Yes/No Questions}.
\newblock \emph{arXiv preprint arXiv:1905.10044}, 2019.

\bibitem[Courbariaux et~al.(2016)Courbariaux, Hubara, Soudry, et~al.]{courbariaux2016binarized}
Matthieu Courbariaux, Itay Hubara, Daniel Soudry, et~al.
\newblock {Binarized Neural Networks: Training Deep Neural Networks with Weights and Activations Constrained to +1 or -1}.
\newblock \emph{arXiv preprint arXiv:1602.02830}, 2016.

\bibitem[Dagan et~al.(2022)Dagan, Roth, Zanzotto, et~al.]{dagan2022recognizing}
Ido Dagan, Dan Roth, Fabio Zanzotto, et~al.
\newblock \emph{{Recognizing Textual Entailment: Models and Applications}}.
\newblock Springer Nature, 2022.

\bibitem[Dao et~al.(2022)Dao, Fu, Ermon, et~al.]{dao2022flashattention}
Tri Dao, Dan Fu, Stefano Ermon, et~al.
\newblock {FlashAttention: Fast and Memory-Efficient Exact Attention with IO-Awareness}.
\newblock \emph{NeurIPS}, 2022.

\bibitem[Darvish~Rouhani et~al.(2020)Darvish~Rouhani, Lo, Zhao, Liu, Fowers, Ovtcharov, Vinogradsky, Massengill, Yang, Bittner, et~al.]{darvish2020pushing}
Bita Darvish~Rouhani, Daniel Lo, Ritchie Zhao, Ming Liu, Jeremy Fowers, Kalin Ovtcharov, Anna Vinogradsky, Sarah Massengill, Lita Yang, Ray Bittner, et~al.
\newblock Pushing the limits of narrow precision inferencing at cloud scale with microsoft floating point.
\newblock \emph{Advances in neural information processing systems}, 33:\penalty0 10271--10281, 2020.

\bibitem[Dettmers and Zettlemoyer(2023)]{dettmers2023case}
Tim Dettmers and Luke Zettlemoyer.
\newblock {The Case for 4-bit Precision: k-bit Inference Scaling Laws}.
\newblock In \emph{ICML}, 2023.

\bibitem[Dettmers et~al.(2022)Dettmers, Lewis, Belkada, et~al.]{dettmers2022gpt3}
Tim Dettmers, Mike Lewis, Younes Belkada, et~al.
\newblock {GPT3.int8(): 8-bit Matrix Multiplication for Transformers at Scale}.
\newblock \emph{NeurIPS}, 2022.

\bibitem[Elhoushi et~al.(2021)Elhoushi, Chen, Shafiq, et~al.]{elhoushi2021deepshift}
Mostafa Elhoushi, Zihao Chen, Farhan Shafiq, et~al.
\newblock {DeepShift: Towards Multiplication-Less Neural Networks}.
\newblock In \emph{CVPR}, 2021.

\bibitem[Frantar and Alistarh(2022)]{frantar2022optimal}
Elias Frantar and Dan Alistarh.
\newblock {Optimal Brain Compression: A Framework for Accurate Post-Training Quantization and Pruning}.
\newblock \emph{NeurIPS}, 2022.

\bibitem[Frantar et~al.(2022)Frantar, Ashkboos, Hoefler, et~al.]{frantar2022optq}
Elias Frantar, Saleh Ashkboos, Torsten Hoefler, et~al.
\newblock {OPTQ: Accurate Quantization for Generative Pre-trained Transformers}.
\newblock In \emph{ICLR}, 2022.

\bibitem[Gholami et~al.(2024)Gholami, Yao, Kim, et~al.]{flops_memory_summary}
Amir Gholami, Zhewei Yao, Sehoon Kim, et~al.
\newblock {AI and Memory Wall}.
\newblock \emph{IEEE Micro Journal}, 2024.

\bibitem[Gromov et~al.(2024)Gromov, Tirumala, Shapourian, et~al.]{gromov2024unreasonable}
Andrey Gromov, Kushal Tirumala, Hassan Shapourian, et~al.
\newblock {The Unreasonable Ineffectiveness of the Deeper Layers}.
\newblock \emph{arXiv preprint arXiv:2403.17887}, 2024.

\bibitem[Guo et~al.(2017)Guo, Yao, Zhao, et~al.]{guo2017network}
Yiwen Guo, Anbang Yao, Hao Zhao, et~al.
\newblock {Network Sketching: Exploiting Binary Structure in Deep CNNs}.
\newblock In \emph{CVPR}, 2017.

\bibitem[Gwee et~al.(2008)Gwee, Chang, Shi, et~al.]{gwee2008low}
Bah-Hwee Gwee, Joseph~S Chang, Yiqiong Shi, et~al.
\newblock {A Low-Voltage Micropower Asynchronous Multiplier With Shift–Add Multiplication Approach}.
\newblock \emph{IEEE Transactions on Circuits and Systems I: Regular Papers}, 2008.

\bibitem[Han et~al.(2016)Han, Liu, Mao, et~al.]{han2016eie}
Song Han, Xingyu Liu, Huizi Mao, et~al.
\newblock {EIE: Efficient Inference Engine on Compressed Deep Neural Network}.
\newblock \emph{ACM SIGARCH Computer Architecture News}, 2016.

\bibitem[Harma et~al.(2024)Harma, Chakraborty, Kostenok, Mishin, Ha, Falsafi, Jaggi, Liu, Oh, Subramanian, and Yazdanbakhsh]{qs-theory}
Simla~Burcu Harma, Ayan Chakraborty, Elizaveta Kostenok, Danila Mishin, Dongho Ha, Babak Falsafi, Martin Jaggi, Ming Liu, Yunho Oh, Suvinay Subramanian, and Amir Yazdanbakhsh.
\newblock {Effective Interplay between Sparsity and Quantization: From Theory to Practice}.
\newblock \emph{arXiv preprint arXiv:2405.20935}, 2024.

\bibitem[Hassibi et~al.(1993)Hassibi, Stork, and Wolff]{hassibi1993optimal}
Babak Hassibi, David~G Stork, and Gregory~J Wolff.
\newblock {Optimal Brain Surgeon and General Network Pruning}.
\newblock In \emph{IEEE international conference on neural networks}, 1993.

\bibitem[Hendrycks et~al.(2020)Hendrycks, Burns, Basart, Zou, Mazeika, Song, and Steinhardt]{hendrycks2020measuring}
Dan Hendrycks, Collin Burns, Steven Basart, Andy Zou, Mantas Mazeika, Dawn Song, and Jacob Steinhardt.
\newblock Measuring massive multitask language understanding.
\newblock \emph{arXiv preprint arXiv:2009.03300}, 2020.

\bibitem[Horowitz(2014)]{horowitz20141}
Mark Horowitz.
\newblock {1.1 Computing's Energy Problem (and what we can do about it)}.
\newblock In \emph{ISSCC}, 2014.

\bibitem[Huang et~al.(2024)Huang, Liu, Qin, et~al.]{huang2024billm}
Wei Huang, Yangdong Liu, Haotong Qin, et~al.
\newblock {BiLLM: Pushing the Limit of Post-Training Quantization for LLMs}.
\newblock \emph{arXiv preprint arXiv:2402.04291}, 2024.

\bibitem[Jeon et~al.(2020)Jeon, Park, Kwon, et~al.]{jeon2020biqgemm}
Yongkweon Jeon, Baeseong Park, Se~Jung Kwon, et~al.
\newblock {BiQGEMM: Matrix Multiplication with Lookup Table For Binary-Coding-based Quantized DNNs}.
\newblock In \emph{SC}, 2020.

\bibitem[Jeon et~al.(2022)Jeon, Lee, Cho, et~al.]{biq}
Yongkweon Jeon, Chungman Lee, Eulrang Cho, et~al.
\newblock {Mr.BiQ: Post-Training Non-Uniform Quantization based on Minimizing the Reconstruction Error}.
\newblock In \emph{CVPR}, 2022.

\bibitem[Jiang et~al.(2023)Jiang, Sablayrolles, Mensch, et~al.]{jiang2023mistral}
Albert~Q Jiang, Alexandre Sablayrolles, Arthur Mensch, et~al.
\newblock {Mistral 7B}.
\newblock \emph{arXiv preprint arXiv:2310.06825}, 2023.

\bibitem[Juefei-Xu et~al.(2017)Juefei-Xu, Naresh~Boddeti, and Savvides]{juefei2017local}
Felix Juefei-Xu, Vishnu Naresh~Boddeti, and Marios Savvides.
\newblock {Local Binary Convolutional Neural Networks}.
\newblock In \emph{CVPR}, 2017.

\bibitem[Kwon et~al.(2021)Kwon, Lee, Jeon, et~al.]{kwon2020post}
Se~Jung Kwon, Dongsoo Lee, Yongkweon Jeon, et~al.
\newblock {Post-Training Weighted Quantization of Neural Networks for Language Models}.
\newblock \url{https://openreview.net/forum?id=2Id6XxTjz7c}, 2021.

\bibitem[Lee et~al.(2023)Lee, Kim, Kwon, and Lee]{lee2023flexround}
Jung~Hyun Lee, Jeonghoon Kim, Se~Jung Kwon, and Dongsoo Lee.
\newblock Flexround: Learnable rounding based on element-wise division for post-training quantization.
\newblock In \emph{International Conference on Machine Learning}, pages 18913--18939. PMLR, 2023.

\bibitem[Lee et~al.(2024)Lee, Kim, Yang, Kwon, Yang, Yoo, and Lee]{lee2024lrq}
Jung~Hyun Lee, Jeonghoon Kim, June~Yong Yang, Se~Jung Kwon, Eunho Yang, Kang~Min Yoo, and Dongsoo Lee.
\newblock Lrq: Optimizing post-training quantization for large language models by learning low-rank weight-scaling matrices.
\newblock \emph{arXiv preprint arXiv:2407.11534}, 2024.

\bibitem[Li et~al.(2023)Li, Liu, Yang, et~al.]{li2023denseshift}
Xinlin Li, Bang Liu, Rui~Heng Yang, et~al.
\newblock {DenseShift: Towards Accurate and Efficient Low-Bit Power-of-Two Quantization}.
\newblock In \emph{ICCV}, 2023.

\bibitem[Li et~al.(2019)Li, Dong, and Wang]{li2019additive}
Yuhang Li, Xin Dong, and Wei Wang.
\newblock {Additive Powers-of-Two Quantization: An Efficient Non-uniform Discretization for Neural Networks}.
\newblock \emph{arXiv preprint arXiv:1909.13144}, 2019.

\bibitem[Lin et~al.(2023)Lin, Tang, Tang, et~al.]{lin2023awq}
Ji~Lin, Jiaming Tang, Haotian Tang, et~al.
\newblock {AWQ: Activation-aware Weight Quantization for LLM Compression and Acceleration}.
\newblock \emph{arXiv preprint arXiv:2306.00978}, 2023.

\bibitem[Liu et~al.(2023)Liu, Oguz, Zhao, et~al.]{liu2023llm}
Zechun Liu, Barlas Oguz, Changsheng Zhao, et~al.
\newblock {LLM-QAT: Data-free Quantization Aware Training for Large Language Models}.
\newblock \emph{arXiv preprint arXiv:2305.17888}, 2023.

\bibitem[Ma et~al.(2023)Ma, Fang, and Wang]{ma2023llm}
Xinyin Ma, Gongfan Fang, and Xinchao Wang.
\newblock {LLM-Pruner: On the Structural Pruning of Large Language Models}.
\newblock \emph{NeurIPS}, 2023.

\bibitem[Merity et~al.(2017)Merity, Xiong, Bradbury, et~al.]{merity2017pointer}
Stephen Merity, Caiming Xiong, James Bradbury, et~al.
\newblock {Pointer Sentinel Mixture Models}.
\newblock In \emph{ICLR}, 2017.

\bibitem[Mesnard et~al.(2024)Mesnard, Hardin, et~al.]{team2024gemma}
Thomas Mesnard, Cassidy Hardin, et~al.
\newblock {Gemma: Open Models Based on Gemini Research and Technology}.
\newblock \emph{arXiv preprint arXiv:2403.08295}, 2024.

\bibitem[Mostafazadeh et~al.(2017)Mostafazadeh, Roth, Louis, et~al.]{mostafazadeh2017lsdsem}
Nasrin Mostafazadeh, Michael Roth, Annie Louis, et~al.
\newblock {LSDSem 2017 Shared Task: The Story Cloze Test}.
\newblock In \emph{Proceedings of the 2nd Workshop on Linking Models of Lexical, Sentential and Discourse-level Semantics}, 2017.

\bibitem[Mozaffari et~al.(2024)Mozaffari, Yazdanbakhsh, Zhang, and Mehri~Dahnavi]{slope}
Mohammad Mozaffari, Amir Yazdanbakhsh, Zhao Zhang, and Maryam Mehri~Dahnavi.
\newblock {SLoPe: Double-Pruned Sparse Plus Lazy Low-rank Adapter Pretraining of LLMs}.
\newblock \emph{arXiv preprint arXiv:2405.16325}, 2024.

\bibitem[{NVIDIA Corporation}(2020)]{nvidia2020a100}
{NVIDIA Corporation}.
\newblock {NVIDIA A100 Tensor Core GPU}.
\newblock \url{https://www.nvidia.com/content/dam/en-zz/Solutions/Data-Center/a100/pdf/nvidia-a100-datasheet-us-nvidia-1758950-r4-web.pdf}, 2020.
\newblock Datasheet.

\bibitem[{OpenAI}(2023)]{chatgpt}
{OpenAI}.
\newblock {ChatGPT: Language Model for Dialogue Generation}.
\newblock \url{https://www.openai.com/chatgpt/}, 2023.
\newblock Website.

\bibitem[OpenAI(2023)]{gpt4}
OpenAI.
\newblock {GPT-4 Technical Report}.
\newblock \emph{arXiv preprint arXiv:2303.08774}, 2023.

\bibitem[Park et~al.(2022)Park, Park, Kim, et~al.]{park2022lut}
Gunho Park, Baeseong Park, Minsub Kim, et~al.
\newblock {LUT-GEMM: Quantized Matrix Multiplication based on LUTs for Efficient Inference in Large-Scale Generative Language Models}.
\newblock \emph{arXiv preprint arXiv:2206.09557}, 2022.

\bibitem[Scao et~al.(2022)Scao, Fan, Akiki, et~al.]{le2023bloom}
Teven~Le Scao, Angela Fan, Christopher Akiki, et~al.
\newblock {BLOOM: A 176B-Parameter Open-Access Multilingual Language Model}.
\newblock \emph{arXiv preprint arXiv:2211.05100}, 2022.

\bibitem[Sentieys(2021)]{sentieys2021approximate}
Olivier Sentieys.
\newblock {Approximate Computing for DNN}.
\newblock In \emph{CSW 2021-HiPEAC Computing Systems Week}, 2021.

\bibitem[Shao et~al.(2023)Shao, Chen, Zhang, Xu, Zhao, Li, Zhang, Gao, Qiao, and Luo]{shao2023omniquant}
Wenqi Shao, Mengzhao Chen, Zhaoyang Zhang, Peng Xu, Lirui Zhao, Zhiqian Li, Kaipeng Zhang, Peng Gao, Yu~Qiao, and Ping Luo.
\newblock Omniquant: Omnidirectionally calibrated quantization for large language models.
\newblock \emph{arXiv preprint arXiv:2308.13137}, 2023.

\bibitem[Shen et~al.(2024)Shen, Kong, Yang, et~al.]{shen2024edgeqat}
Xuan Shen, Zhenglun Kong, Changdi Yang, et~al.
\newblock {EdgeQAT: Entropy and Distribution Guided Quantization-Aware Training for the Acceleration of Lightweight LLMs on the Edge}.
\newblock \emph{arXiv preprint arXiv:2402.10787}, 2024.

\bibitem[Shi et~al.(2022)Shi, You, Zhao, Wang, and Lin]{shi2022nasa}
Huihong Shi, Haoran You, Yang Zhao, Zhongfeng Wang, and Yingyan Lin.
\newblock Nasa: Neural architecture search and acceleration for hardware inspired hybrid networks.
\newblock In \emph{Proceedings of the 41st IEEE/ACM International Conference on Computer-Aided Design}, pages 1--9, 2022.

\bibitem[Shu et~al.(2021)Shu, Wang, Chen, et~al.]{shu2021adder}
Han Shu, Jiahao Wang, Hanting Chen, et~al.
\newblock {Adder Attention for Vision Transformer}.
\newblock \emph{NeurIPS}, 2021.

\bibitem[Sun et~al.(2023)Sun, Liu, Bair, et~al.]{sun2023simple}
Mingjie Sun, Zhuang Liu, Anna Bair, et~al.
\newblock {A Simple and Effective Pruning Approach for Large Language Models}.
\newblock \emph{arXiv preprint arXiv:2306.11695}, 2023.

\bibitem[Tata and Patel(2003)]{tata2003piqa}
Sandeep Tata and Jignesh~M Patel.
\newblock {PiQA: An Algebra for Querying Protein Data Sets}.
\newblock In \emph{SSDBM}, 2003.

\bibitem[Touvron et~al.(2023{\natexlab{a}})Touvron, Lavril, Izacard, et~al.]{touvron2023llama}
Hugo Touvron, Thibaut Lavril, Gautier Izacard, et~al.
\newblock {LLaMA: Open and Efficient Foundation Language Models}.
\newblock \emph{arXiv preprint arXiv:2302.13971}, 2023{\natexlab{a}}.

\bibitem[Touvron et~al.(2023{\natexlab{b}})Touvron, Martin, Stone, et~al.]{touvron2023llama2}
Hugo Touvron, Louis Martin, Kevin Stone, et~al.
\newblock {Llama 2: Open Foundation and Fine-Tuned Chat Models}.
\newblock \emph{arXiv preprint arXiv:2307.09288}, 2023{\natexlab{b}}.

\bibitem[Waisberg et~al.(2023)Waisberg, Ong, Masalkhi, et~al.]{bard}
Ethan Waisberg, Joshua Ong, Mouayad Masalkhi, et~al.
\newblock {Google’s AI chatbot “Bard”: A Side-by-Side Comparison with ChatGPT and its Utilization in Ophthalmology}.
\newblock \emph{Eye}, 2023.

\bibitem[Wang et~al.(2022)Wang, Zhao, Tang, et~al.]{wang2022shift}
Guangting Wang, Yucheng Zhao, Chuanxin Tang, et~al.
\newblock {When Shift Operation Meets Vision Transformer: An Extremely Simple Alternative to Attention Mechanism}.
\newblock In \emph{AAAI}, 2022.

\bibitem[Wang et~al.(2021)Wang, Huang, Han, et~al.]{adder_hardware}
Yunhe Wang, Mingqiang Huang, Kai Han, et~al.
\newblock {AdderNet and its Minimalist Hardware Design for Energy-Efficient Artificial Intelligence}.
\newblock \emph{arXiv preprint arXiv:2101.10015}, 2021.

\bibitem[Wu et~al.(2018)Wu, Wan, Yue, et~al.]{wu2018shift}
Bichen Wu, Alvin Wan, Xiangyu Yue, et~al.
\newblock {Shift: A Zero FLOP, Zero Parameter Alternative to Spatial Convolutions}.
\newblock In \emph{CVPR}, 2018.

\bibitem[Xiao et~al.(2023)Xiao, Lin, Seznec, et~al.]{xiao2023smoothquant}
Guangxuan Xiao, Ji~Lin, Mickael Seznec, et~al.
\newblock {SmoothQuant: Accurate and Efficient Post-Training Quantization for Large Language Models}.
\newblock In \emph{ICML}, 2023.

\bibitem[Xu et~al.(2018)Xu, Yao, Lin, et~al.]{xu2018alternating}
Chen Xu, Jianqiang Yao, Zhouchen Lin, et~al.
\newblock {Alternating Multi-bit Quantization for Recurrent Neural Networks}.
\newblock \emph{arXiv preprint arXiv:1802.00150}, 2018.

\bibitem[Xu et~al.(2020)Xu, Xu, Chen, et~al.]{adder_distillation}
Yixing Xu, Chang Xu, Xinghao Chen, et~al.
\newblock {Kernel Based Progressive Distillation for Adder Neural Networks}.
\newblock In \emph{NeurIPS}, 2020.

\bibitem[Xue and Liu(1986)]{xue1986adaptive}
Ping Xue and Bede Liu.
\newblock {Adaptive Equalizer Based on a Power-Of-Two-Quantized-LMF Algorithm}.
\newblock \emph{IEEE transactions on acoustics, speech, and signal processing}, 1986.

\bibitem[Yang et~al.(2023)Yang, Wang, Shen, et~al.]{yang2023gated}
Songlin Yang, Bailin Wang, Yikang Shen, et~al.
\newblock {Gated Linear Attention Transformers with Hardware-Efficient Training}.
\newblock \emph{arXiv preprint arXiv:2312.06635}, 2023.

\bibitem[Yao et~al.(2022)Yao, Yazdani~Aminabadi, Zhang, et~al.]{yao2022zeroquant}
Zhewei Yao, Reza Yazdani~Aminabadi, Minjia Zhang, et~al.
\newblock {ZeroQuant: Efficient and Affordable Post-Training Quantization for Large-Scale Transformers}.
\newblock \emph{NeurIPS}, 2022.

\bibitem[You et~al.(2020)You, Chen, Zhang, et~al.]{you2020shiftaddnet}
Haoran You, Xiaohan Chen, Yongan Zhang, et~al.
\newblock {ShiftAddNet: A Hardware-Inspired Deep Network}.
\newblock \emph{NeurIPS}, 2020.

\bibitem[You et~al.(2022)You, Li, Huihong, et~al.]{you2022shiftaddnas}
Haoran You, Baopu Li, Shi Huihong, et~al.
\newblock {ShiftAddNAS: Hardware-Inspired Search for More Accurate and Efficient Neural Networks}.
\newblock In \emph{ICLR}, 2022.

\bibitem[You et~al.(2024{\natexlab{a}})You, Fu, Wang, et~al.]{you2024linear}
Haoran You, Yichao Fu, Zheng Wang, et~al.
\newblock {When Linear Attention Meets Autoregressive Decoding: Towards More Effective and Efficient Linearized Large Language Models}.
\newblock In \emph{ICML}, 2024{\natexlab{a}}.

\bibitem[You et~al.(2024{\natexlab{b}})You, Shi, Guo, et~al.]{you2024shiftaddvit}
Haoran You, Huihong Shi, Yipin Guo, et~al.
\newblock {ShiftAddViT: Mixture of multiplication primitives towards efficient vision transformer}.
\newblock \emph{NeurIPS}, 2024{\natexlab{b}}.

\bibitem[Yuan et~al.(2024)Yuan, Shang, Zhou, Dong, Xue, Wu, Li, Gu, Lee, Yan, et~al.]{yuan2024llm}
Zhihang Yuan, Yuzhang Shang, Yang Zhou, Zhen Dong, Chenhao Xue, Bingzhe Wu, Zhikai Li, Qingyi Gu, Yong~Jae Lee, Yan Yan, et~al.
\newblock Llm inference unveiled: Survey and roofline model insights.
\newblock \emph{arXiv preprint arXiv:2402.16363}, 2024.

\bibitem[Zhang et~al.(2022)Zhang, Roller, Goyal, et~al.]{zhang2022opt}
Susan Zhang, Stephen Roller, Naman Goyal, et~al.
\newblock {OPT: Open Pre-trained Transformer Language Models}.
\newblock \emph{arXiv preprint arXiv:2205.01068}, 2022.

\bibitem[Zhao et~al.(2020)Zhao, Li, Wang, et~al.]{zhao2020dnn}
Yang Zhao, Chaojian Li, Yue Wang, et~al.
\newblock {DNN-Chip Predictor: An Analytical Performance Predictor for DNN Accelerators with Various Dataflows and Hardware Architectures}.
\newblock In \emph{ICASSP}, 2020.

\bibitem[Zhu et~al.(2024)Zhu, Zhang, Sifferman, Sheaves, Wang, Richmond, Zhou, and Eshraghian]{zhu2024scalable}
Rui-Jie Zhu, Yu~Zhang, Ethan Sifferman, Tyler Sheaves, Yiqiao Wang, Dustin Richmond, Peng Zhou, and Jason~K Eshraghian.
\newblock Scalable matmul-free language modeling.
\newblock \emph{arXiv preprint arXiv:2406.02528}, 2024.

\end{thebibliography}
}

\newpage
\appendix

\section{Complete Accuracy \& Latency \& Energy Data for OPT Models}
\label{sec:opt_appendix}

We supply the complete quantitative accuracy, latency, and energy data measured on the OPT model family in Tab.~\ref{tab:opt_ppl_appendix}, \ref{tab:latency_opt_appendix}, and \ref{tab:energy_opt}, respectively.

\begin{table}[h]
\centering
\renewcommand{\arraystretch}{1.2}
\caption{Perplexity comparisons of the OPT models on WikiText-2. Note that we set the group size of all methods as the number of columns following the setting of OPTQ~\cite{frantar2022optq} for a fair comparison.}
\resizebox{\linewidth}{!}{
\begin{tabular}{l|c|cccccccccc}
\Xhline{3\arrayrulewidth}
\textbf{OPT (PPL $\downarrow$)} & \textbf{Bits} & \textbf{125M} & \textbf{350M} & \textbf{1.3B} & \textbf{2.7B} & \textbf{6.7B} & \textbf{13B} & \textbf{30B} & \textbf{66B} \\
\Xhline{3\arrayrulewidth}
FP16 & 16 & 27.65 & 22.00 & 14.62 & 12.47 & 10.86 & 10.13 & 9.56 & 9.34 \\
\hline
OPTQ~\cite{frantar2022optq} & 3 & 53.85 & 33.79 & 20.97 & 16.88 & 14.86 & 11.61 & 10.27 & 14.16 \\
LUT-GEMM~\cite{park2022lut} & 3 & 60.00 & 42.32 & 49.10 & 17.55 & 17.44 & 12.50 & 139.90 & 100.33 \\
AWQ~\cite{lin2023awq} & 3 & 54.75 & 35416.00 & 24.60 & 39.01 & 16.47 & 16.53 & 31.01 & 5622.00 \\
\textbf{Ours (Lat.)} & 3 & 56.96 & 28.72 & 19.69 & 15.28 & 11.80 & 10.70 & 9.89 & 9.62 \\
\Xhline{3\arrayrulewidth}
OPTQ~\cite{frantar2022optq} & 2 & 2467.50 & 10433.30 & 4737.05 & 6294.68 & 442.63 & 126.09 & 71.70 & 20.91 \\
LUT-GEMM~\cite{park2022lut} & 2 & 4844.32 & 2042.90 & 3851.50 & 616.30 & 17455.52 & 4963.27 & 7727.27 & 6246.00 \\
AWQ~\cite{lin2023awq} & 2 & 3514.61 & 18313.24 & 9472.81 & 22857.70 & 8168.30 & 5014.92 & 7780.96 & 103843.84 \\
\textbf{Ours (Lat.)} & 2 & 712.55 & 445.78 & 40.28 & 50.95 & 18.56 & 14.76 & 12.55 & 12.20 \\
\hline
\textbf{Ours (Mixed)} & 2.2 & 435.84 & 279.19 & 27.37 & 31.97 & 17.99 & 13.79 &  11.62 &  11.17 \\
\Xhline{3\arrayrulewidth}
\end{tabular}
}
\label{tab:opt_ppl_appendix}
\end{table}

\begin{table}[h]
\centering
\renewcommand{\arraystretch}{1.2}
\caption{A100 GPU latency comparisons on the OPT model family. 
}
\resizebox{0.85\linewidth}{!}{
\begin{tabular}{l|c|cccccccc}
\Xhline{3\arrayrulewidth}
\textbf{OPT Latency (ms)} & \textbf{Bits} & \textbf{125M} & \textbf{350M} & \textbf{1.3B} & \textbf{2.7B} & \textbf{6.7B} & \textbf{13B} & \textbf{30B} & \textbf{66B} \\
\Xhline{3\arrayrulewidth}
FP16 & 16 & 7.8 & 15.1 & 16.7 & 20.9 & 22.2 & 29.5 & 51.7 & OOM \\
\hline
OPTQ~\cite{frantar2022optq} & 3 & 8.3 & 15.9 & 15.0 & 21.5 & 21.1 & 26.4 & 30.1 & 51.5 \\
LUT-GEMM~\cite{park2022lut} & 3 & 6.3 & 11.7 & 12.6 & 15.5 & 17.0 & 19.5 & 23.7 & 39.5 \\
AWQ~\cite{lin2023awq} & 3 & 6.2 & 12.1 & 12.3 & 16.3 & 16.3 & 20.0 & 24.5 & 40.9 \\
\textbf{Ours (Lat.)} & 3 & 6.4 & 13.3 & 12.6 & 16.6 & 16.9 & 20.8 & 30.7 & 54.1 \\
\hline
OPTQ~\cite{frantar2022optq} & 2 & 8.2 & 16.1 & 15.9 & 19.7 & 19.9 & 24.7 & 31.5 & 50.4 \\
LUT-GEMM~\cite{park2022lut} & 2 & 6.2 & 11.8 & 11.8 & 15.7 & 15.7 & 19.8 & 23.6 & 33.2 \\
AWQ~\cite{lin2023awq} & 2 & 6.2 & 12.1 & 12.3 & 16.3 & 16.3 & 20.0 & 24.5 & 40.9 \\
\textbf{Ours (Lat.)} & 2 & 6.3 & 12.4 & 12.3 & 16.9 & 16.9 & 20.9 & 25.4 & 42.9 \\
\hline
\textbf{Ours (Mixed)} & 2.2 & 6.3 & 12.6 & 12.5 & 16.8 & 16.7 & 21.0 & 27.1 & 45.7 \\
\Xhline{3\arrayrulewidth}
\end{tabular}
}
\label{tab:latency_opt_appendix}
\leftline{\small \quad\quad\quad\quad * Note that we use AWQ's open-sourced \texttt{INT4} kernel for measuring its latency.}
\end{table}

\begin{table}[h]
\centering
\renewcommand{\arraystretch}{1.2}
\caption{Energy comparisons on the OPT model family.}
\resizebox{\linewidth}{!}{
\begin{tabular}{l|c|cccccccc}
\Xhline{3\arrayrulewidth}
\textbf{OPT Energy (J)} & \textbf{Bits} & \textbf{125M} & \textbf{350M} & \textbf{1.3B} & \textbf{2.7B} & \textbf{6.7B} & \textbf{13B} & \textbf{30B} & \textbf{66B} \\
\Xhline{3\arrayrulewidth}
FP16 & 16 & 29.26 & 83.72 & 310.33 & 625.80 & 1573.41 & 3036.99 & 7088.39 & 15539.87 \\
\hline
OPTQ~\cite{frantar2022optq} & 3 & 13.77 & 28.63 & 90.12 & 167.37 & 399.28 & 745.37 & 1695.45 & 3658.17\\
LUT-GEMM~\cite{park2022lut} & 3 & 12.83 & 25.19 & 75.73 & 137.08 & 321.54 & 592.31 & 1332.95 & 2858.87\\
\textbf{Ours (Lat.)} & 3 & 11.68 & 21.13 & 59.59 & 103.53 & 235.45 & 424.58 & 938.98 & 1990.17\\
\hline
OPTQ~\cite{frantar2022optq} & 2 & 12.58 & 24.42 & 73.27 & 132.30 & 309.50 & 570.06 & 1283.04 & 2749.32\\
LUT-GEMM~\cite{park2022lut} & 2 & 12.48 & 23.96 & 70.90 & 127.07 & 295.77 & 542.28 & 1215.36 & 2599.77\\
\textbf{Ours (Lat.)} & 2 & 11.41 & 20.17 & 55.80 & 95.67 & 215.27 & 385.31 & 846.54 & 1786.62\\
\hline
\textbf{Ours (Mixed)} & 2.2 & 11.45 & 20.33 & 56.43 & 96.98 & 218.64 & 391.86 & 861.95 & 1820.55\\
\Xhline{3\arrayrulewidth}
\end{tabular}
}
\label{tab:energy_opt}
\end{table}

\newpage
\section{Complete Accuracy \& Latency \& Energy Data for LLaMA Models}
\label{sec:llama_appendix}

We supply the complete quantitative accuracy, latency, and energy data measured on the LLaMA model family in Tab.~\ref{tab:llama_ppl_appendix}, \ref{tab:llama_lat_appendix}, and \ref{tab:llama_energy_appendix}, respectively.

\begin{table}[h]
\centering
\setlength{\tabcolsep}{4pt}
\renewcommand{\arraystretch}{1.2}
\caption{Perplexity comparisons of the LLaMA models on WikiText-2.}
\resizebox{0.65\linewidth}{!}{
\begin{tabular}{l|c|x{1cm}x{1cm}x{1cm}|x{1cm}x{1cm}}
\Xhline{3\arrayrulewidth}
\multirow{2}[1]{*}{\textbf{LLaMA (PPL $\downarrow$)}} & \multirow{2}[1]{*}{\textbf{Bits}} & \multicolumn{3}{c|}{\textbf{LLaMA-2}} & \multicolumn{2}{c}{\textbf{LLaMA-3}} \\
\cline{3-7}      &  & \textbf{7B} & \textbf{13B} & \textbf{70B} & \textbf{8B} & \textbf{70B} \\
\Xhline{3\arrayrulewidth}
FP16 & 16 & 5.47 & 4.88 & 3.32 & 6.14 & 2.86 \\
\hline
OPTQ~\cite{frantar2022optq} & 3 & 6.43 & 5.48 & 3.88 & 13.69 & 4.91 \\
LUT-GEMM~\cite{park2022lut} & 3 & 7.02 & 5.89 & 4.01 & 11.10 & 5.92 \\
AWQ~\cite{lin2023awq} & 3 & 6.24 & 5.32 & 3.74 & 8.15 & 4.69 \\
\textbf{Ours (Lat.)} & 3 &6.04 & 5.33 & 3.72 & 7.71 & 4.66\\
\Xhline{3\arrayrulewidth}
OPTQ~\cite{frantar2022optq} & 2 & 19.92 & 12.75 & 6.82 & 398.0 & 26.65 \\
LUT-GEMM~\cite{park2022lut} & 2 & 2242.0 & 2791.0 & 136.4 & 19096 & 3121 \\
AWQ~\cite{lin2023awq} & 2 & 2.22e5 & 1.22e5& 7.24e4 & 1.71e6 & 1.72e6 \\
\textbf{Ours (Lat.)} & 2 & 9.58 & 12.57& 5.71& 34.4& 12.4\\
\Xhline{3\arrayrulewidth}
\end{tabular}
}
\label{tab:llama_ppl_appendix}
\leftline{\small \quad\quad\quad\quad\quad\quad\quad\quad * Note that the group size is set to 128 following \cite{park2022lut,lin2023awq}.}
\end{table}

\begin{table}[h]
\centering
\setlength{\tabcolsep}{3.5pt}
\renewcommand{\arraystretch}{1.2}
\caption{A100 GPU latency comparisons of the LLaMA models.}
\resizebox{0.65\linewidth}{!}{
\begin{tabular}{l|c|x{1cm}x{1cm}x{1cm}|x{1cm}x{0.9cm}}
\Xhline{3\arrayrulewidth}
\multirow{2}[1]{*}{\textbf{LLaMA Latency (ms)}} & \multirow{2}[1]{*}{\textbf{Bits}} &  \multicolumn{3}{c|}{\textbf{LLaMA-2}} & \multicolumn{2}{c}{\textbf{LLaMA-3}} \\
\cline{3-7}      &   & \textbf{7B} & \textbf{13B} & \textbf{70B} & \textbf{8B} & \textbf{70B} \\
\Xhline{3\arrayrulewidth}
FP16 & 16 &  32.6 & 43.1 & OOM & 38.8 & OOM \\
\hline
OPTQ~\cite{frantar2022optq} & 3 &  31.1 & 42.2 & 81.9 & 36.2 & 90.7 \\
LUT-GEMM~\cite{park2022lut} & 3 &  27.4 & 34.7 & 72.6 & 31.7 & 77.5 \\
AWQ~\cite{lin2023awq} & 3 &  25.4 & 31.8 & 68.0 & 28.5 & 67.7 \\
\textbf{Ours (Lat.)} & 3 & 26.7 & 33.8 & 70.9 & 31.4 & 72.9 \\
\Xhline{3\arrayrulewidth}
OPTQ~\cite{frantar2022optq} & 2 & 34.2 & 38.8 & 82.5 & 36.8 & 91.2 \\
LUT-GEMM~\cite{park2022lut} & 2 & 27.5 & 33.3 & 71.0 & 31.7 & 77.2 \\
AWQ~\cite{lin2023awq} & 2 & 25.4 & 31.8 & 68.0 & 28.5 & 67.7 \\
\textbf{Ours (Lat.)} & 2 &  27.7 & 33.9 & 72.1 & 31.9 & 78.3 \\
\hline
\textbf{Ours (Mixed)} & 2.2 &  27.2 & 34.3 & 75.1 & 30.1 & 76.4 \\
\Xhline{3\arrayrulewidth}
\end{tabular}
}
\label{tab:llama_lat_appendix}
\end{table}

\begin{table}[h]
\centering
\setlength{\tabcolsep}{4pt}
\renewcommand{\arraystretch}{1.2}
\caption{Energy comparisons of the LLaMA models.}
\resizebox{0.75\linewidth}{!}{
\begin{tabular}{l|c|x{1.2cm}x{1.2cm}x{1.5cm}|x{1.2cm}x{1.5cm}}
\Xhline{3\arrayrulewidth}
\multirow{2}[1]{*}{\textbf{LLaMA Energy (J)}} & \multirow{2}[1]{*}{\textbf{Bits}} & \multicolumn{3}{c|}{\textbf{LLaMA-2}} & \multicolumn{2}{c}{\textbf{LLaMA-3}} \\
\cline{3-7}      &   & \textbf{7B} & \textbf{13B} & \textbf{70B} & \textbf{8B} & \textbf{70B} \\
\Xhline{3\arrayrulewidth}
FP16 & 16 & 1563.44 & 3040.26 & 18482.5 & 1776.05 & 16445.98\\
\hline
OPTQ~\cite{frantar2022optq} & 3 & 383.40 & 728.98 & 4297.33 & 504.07 & 3972.72\\
LUT-GEMM~\cite{park2022lut} & 3 & 305.06 & 574.71 & 3349.01 & 419.64 & 3139.34\\
\textbf{Ours (Lat.)} & 3 & 218.59 & 405.53 & 2309.87 & 326.47 & 2225.71 \\
\Xhline{3\arrayrulewidth}
OPTQ~\cite{frantar2022optq} & 2 & 293.15 & 552.20 & 3212.56 & 406.81 & 3018.87\\
LUT-GEMM~\cite{park2022lut} & 2 & 279.20 & 524.16 & 3037.94 & 391.74 & 2865.81\\
\textbf{Ours (Lat.)} & 2 & 198.33 & 365.85 & 2065.90 & 304.59 & 2011.15 \\
\hline
\textbf{Ours (Mixed)} & 2.2 & 201.69 & 372.40 & 2099.53 & 306.69 & 2066.64 \\
\Xhline{3\arrayrulewidth}
\end{tabular}
}
\label{tab:llama_energy_appendix}
\end{table}

\newpage
\section{Ablation Studies on Multi-Objective Optimization}
\label{sec:ablation_appendix}

We conduct ablation studies on different optimization objectives. As shown in Tab.~\ref{tab:ablation_objective}, our multi-objective optimization demonstrates superior performance in both column-wise and block-wise scaling factor formats. It achieves average perplexity reductions of \hr{123.25, 2.22, and 403.18} compared to the weight-only objective, activation-only objective, and the vanilla combination of both weight and activation objectives, respectively. These experiments validate the effectiveness of our multi-objective optimization approach.

\begin{table}[h]
\centering
\setlength{\tabcolsep}{3.6pt}
\renewcommand{\arraystretch}{1.2}
\caption{Ablation studies on various optimization objectives.}
\resizebox{0.4\linewidth}{!}{
\begin{tabular}{l|ccc}
\Xhline{3\arrayrulewidth}
\textbf{OPT PPL} & \textbf{13B} & \textbf{30B} & \textbf{66B} \\
\Xhline{3\arrayrulewidth}
Wei. Obj. & 13.8 & 222.6 & 163.2 \\
Act. Obj. & 11.7 & 10.5 & 14.3 \\
Wei. + Act. & 45.0 & 16.3 & 1178.1 \\
\hline
\textbf{Ours (Col.-wise)} & \textbf{10.4} & \textbf{9.6} & \textbf{9.4} \\
\textbf{Ours (Blk.-wise)} & \textbf{10.8} & \textbf{9.9} & \textbf{9.6} \\
\Xhline{3\arrayrulewidth}
\end{tabular}
}
\label{tab:ablation_objective}
\end{table}

\section{\hry{Impact of Batch Sizes on Throughput}}
\label{sec:batch_size}

\begin{figure}[h]
    \centering
    \includegraphics[width=0.5\linewidth]{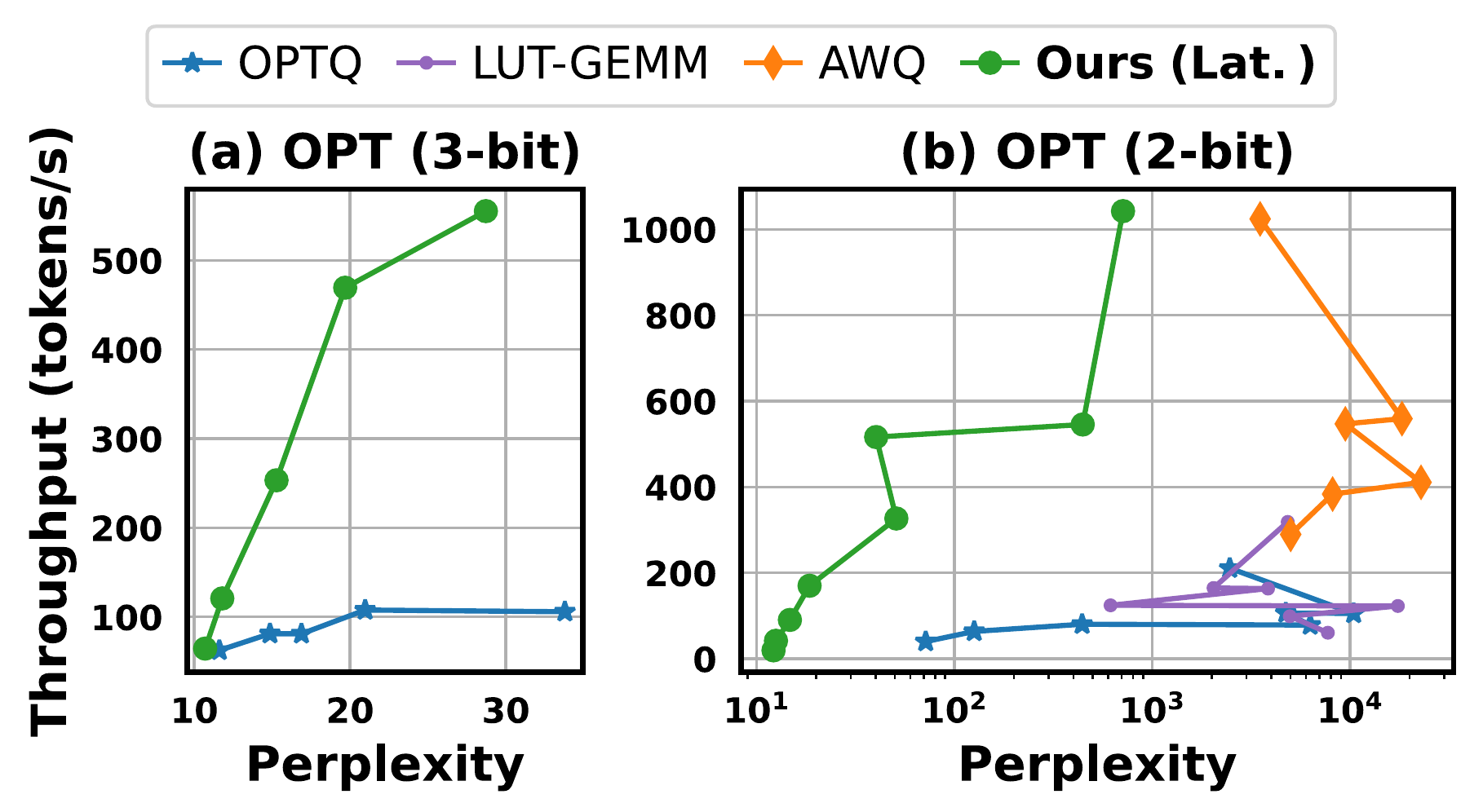}
    \includegraphics[width=\linewidth]{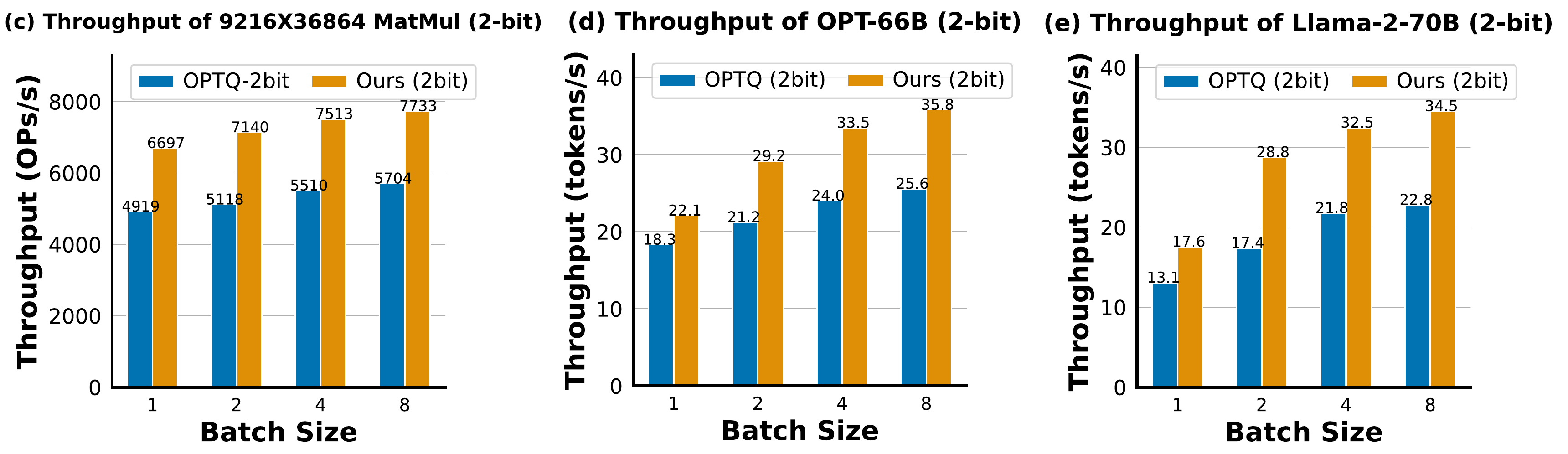}
    \caption{(a-b): Accuracy-throughput tradeoff comparisons among ShiftAddLLM, OPTQ, LUT-GEMM, and AWQ at a batch size of 8. (c) Kernel throughput evaluation under batch sizes of 1, 2, 4, and 8. (d) LLaMA-2-70B end-to-end model throughput evaluation under batch sizes of 1, 2, 4, and 8. (e) OPT-66B end-to-end model throughput evaluation under batch sizes of 1, 2, 4, and 8.}
    \label{fig:batch_size}
\end{figure}

To investigate the impact of batch sizes on the achievable throughput, we have further tested the throughput of our CUDA kernels and end-to-end models with increased batch sizes, as demonstrated in Fig.~\ref{fig:batch_size}. Our ShiftAddLLM still outperforms all three baselines at a batch size of 8 in terms of accuracy-efficiency trade-offs, achieving on average 3.37$\times$/2.55$\times$/1.39$\times$ throughput improvements compared to OPTQ, AWQ, and LUT-GEMM at similar or much better accuracy.

Previously, we assumed a batch size of one for mobile applications where only one user is using the LLM. This assumption also stems from the sequential nature of LLMs during generation, i.e., generating one token at a time based on all previously generated contexts. The assumption of a batch size of 1 is also used in previous literature, such as AWQ, OPTQ, and LUT-GEMM, to measure the latency or throughput for LLM serving.

\section{\hry{Benchmark with More Recent Baselines}}
\label{sec:benchmark}

We further compare our ShiftAddLLM with recent LLM quantization baselines FlexRound~\cite{lee2023flexround} and OmniQuant~\cite{shao2023omniquant} on OPT and LLaMA models.
As shown in Tabs.~\ref{tab:more_baseline_opt} \& \ref{tab:more_baseline_llama}, our ShiftAddLLM consistently shows better accuracy-efficiency trade-offs, achieving an average of 0.15 (4-bit) / 0.39 (3-bit) and 0.30 (4-bit) / 0.52 (3-bit) perplexity reduction, as compared to FlexRound and OmniQuant, respectively. Note that the baseline results are directly obtained from the original paper and follow-up work LRQ~\cite{lee2024lrq}. In addition, we tested OmniQuant at 2 bits ourselves and found it fails for OPT models, whereas ours performs well for OPT models and also achieves an average 1.96 perplexity reduction than OmniQuant on LLaMA at 2 bits.

\begin{table}[h]
  \centering
  \setlength{\tabcolsep}{3.6pt}
  \renewcommand{\arraystretch}{1.2}
  \caption{Perplexity comparisons between ShiftAddLLM and OmniQuant using OPT models and LLaMA models on WikiText-2. The group size is set as the length of rows for OPT models and 128 for LLaMA models following baselines.}
  \resizebox{0.93\linewidth}{!}{
    \begin{tabular}{l|c|cccccccc|ccc}
    \Xhline{3\arrayrulewidth}
    \multirow{2}{*}{\textbf{Method}} & 
    \multirow{2}{*}{\textbf{Bits}} & 
    \multicolumn{8}{c|}{\textbf{OPT}} & 
    \multicolumn{3}{c}{\textbf{LLaMA-2}}
    \\
    \cline{3-10} \cline{11-13} 
     & & \textbf{125M} & \textbf{350M} & \textbf{1.3B} & \textbf{2.7B} & \textbf{6.7B} & \textbf{13B} & \textbf{30B} & \textbf{66B} & 
     \textbf{7B} & \textbf{13B} & \textbf{70B} \\
    \Xhline{3\arrayrulewidth}
    
    OmniQuant~\cite{shao2023omniquant} & 4 & 29.45 & 23.19 & 15.04 & 12.76 & 11.03 & 10.30 & 9.65 & - & 5.58 & 4.95 & - \\
    \textbf{Ours (Acc.)} & 4 & \textbf{28.72} & \textbf{21.59} & \textbf{14.98} & \textbf{12.65} & \textbf{10.95} & \textbf{10.20} & \textbf{9.63} & -  & 5.58 & 4.96 & - \\
    
    \hline
    
    OmniQuant~\cite{shao2023omniquant} & 3 & 35.66 & 28.2 & 16.68 & 13.8 & 11.65 & 10.87 & 10.00 & 9.83 & 6.03 &  5.28 & 3.78\\
    \textbf{Ours (Acc.)} & 3 & \textbf{31.29} & \textbf{24.24} & \textbf{21.53} & \textbf{13.68} & \textbf{11.18} & \textbf{10.39} & \textbf{9.63} & \textbf{9.43} & \textbf{5.89} & \textbf{5.16} & \textbf{3.64} \\
    
    \hline
    
    OmniQuant~\cite{shao2023omniquant} & 2 & 311.39 & 186.9 & 484.51 & 1.1e6 & 9.6e5 & 3.6e4 & 9.3e3 & 5.2e3 & 11.06 & 8.26 & 6.55\\
    \textbf{Ours (Acc.)} & 2 & \textbf{51.15} & \textbf{40.24} & \textbf{29.03} & \textbf{20.78} & \textbf{13.78} & \textbf{12.17} & \textbf{10.67} & \textbf{10.33} & \textbf{8.51} & \textbf{6.77} & \textbf{4.72} \\
    
    \Xhline{3\arrayrulewidth}
    \end{tabular}}
    \label{tab:more_baseline_opt}
\end{table}

\begin{table}[h]
  \centering
  \setlength{\tabcolsep}{3.6pt}
  \renewcommand{\arraystretch}{1.2}
  \caption{Perplexity comparisons between ShiftAddLLM and FlexRound. The group size of FlexRound is set as the length of rows following the paper.}
  \resizebox{0.4\linewidth}{!}{
    \begin{tabular}{l|c|ccc}
    \Xhline{3\arrayrulewidth}
    
    \multirow{2}{*}{\textbf{Method}} & 
    \multirow{2}{*}{\textbf{Bits}} & 
    \multicolumn{3}{c}{\textbf{LLaMA-2}}
    \\
    \cline{3-5}
    & & \textbf{7B} & \textbf{13B} & \textbf{70B}\\
    \Xhline{3\arrayrulewidth}
    
    FlexRound~\cite{lee2023flexround} & 4 & 5.83 & 5.01 & - \\
    \textbf{Ours (Acc.)} & 4 & 5.58 & 4.96 & - \\
    
    \hline
    
    FlexRound~\cite{lee2023flexround} & 3 & 6.34 & 5.59 & 3.92\\
    \textbf{Ours (Acc.)} & 3 & \textbf{5.89} & \textbf{5.16} & \textbf{3.64} \\

    \Xhline{3\arrayrulewidth}
    \end{tabular}
  }
  \label{tab:more_baseline_llama}
\end{table}%

\section{\hry{More Results for Mixed Bit Allocation}}
\label{sec:more_mixed_bit}

To validate the effectiveness and applicability of our automated bit allocation across different LLM models, we evaluated and compared \textit{Ours (Mixed)} with \textit{Ours (Lat.)}. The results are shown in Tab.~\ref{tab:more_mixed_bit}. \textit{Ours (Mixed)} further reduces perplexity by an average of 96.86, 3.23, and 2.63 for OPT, LLaMA, and Gemma models, respectively, under comparable or even less latency. This set of experiments further validates the applicability of our automated bit allocation strategy to different LLMs.

\begin{table}[h]
    \centering
    \setlength{\tabcolsep}{3.6pt}
    \renewcommand{\arraystretch}{1.2}
    \caption{Perplexity and correlation results of our mixed bit allocation.}
    \resizebox{0.8\linewidth}{!}{
    \begin{tabular}{l|c|ccc|ccc|c}
        \Xhline{3\arrayrulewidth}
        \multirow{2}[1]{*}{\textbf{Methods}} & \multirow{2}[1]{*}{\textbf{Bits}} & 
        \multicolumn{3}{c|}{\textbf{OPT}} &
        \multicolumn{3}{c|}{\textbf{LLaMA}} & \multicolumn{1}{c}{\textbf{Gemma}} \\
        \cline{3-9}
        & & \textbf{125M} & \textbf{1.3B} & \textbf{13B} &
        \textbf{2-7B} & \textbf{2-13B} & \textbf{3-8B} & \textbf{2B} \\
        \Xhline{3\arrayrulewidth}
        \multicolumn{2}{c|}{\textbf{Correlation ($\tau$)}} & 0.910 & 0.905 & 0.915 & 0.931 & 0.929 & 0.897 & - \\
        \hline
        \textbf{Ours (Lat.)} & 2 & 712.55 & 40.28 & 14.76 & 9.58 & 12.57 & 34.40 & 16.52 (3 bits) \\
        \textbf{Ours (Mixed)} & 2.2 & 435.84 & 27.37 & 13.79 & 8.97 & 8.16 & 29.72 & 13.89 \\
        \Xhline{3\arrayrulewidth}
    \end{tabular}
    }
    \label{tab:more_mixed_bit}
\end{table}

In addition, we want to clarify that, for each model, we search for the optimal bit allocation with negligible overhead (e.g., 1\%~10\% of the reparameterization time). For example, it takes 0.5 seconds for searching versus 72 seconds for reparameterizing OPT-125M with a single bit configuration, and 1 minute for searching versus 13 minutes for reparameterizing OPT-13B with a single bit configuration. This is achieved by leveraging the proposed proxy criteria (as shown in Sec.~\ref{sec:shiftaddllm-3}), instead of searching according to the reparameterization errors, which is time-consuming and requires running models at each bit. Using the proxy criteria, the bit allocation candidate rankings are highly correlated with the rankings obtained using actual reparameterization errors, with a Kendall $\tau$ of 0.910/0.905/0.915 for OPT-125M/1.3B/13B and 0.931/0.929/0.897 for LLaMA-7B/13B/8B, respectively.

\section{\hry{4-Bit Results and Explanation for Using Lower Bit Widths}}
\label{sec:4-bit}

We further provide the 4-bit results in Tab.~\ref{tab:4bit}. These results show that ShiftAddLLM consistently outperforms the baselines at 4 bits, achieving average perplexity reductions of 0.90/1.32/1.00 and 0.44/0.22/0.02 as compared to OPTQ/LUT-GEMM/AWQ, using OPT models and LLaMA models, respectively.

\begin{table}[h]
\centering
\setlength{\tabcolsep}{3.6pt}
\renewcommand{\arraystretch}{1.2}
\caption{Perplexity comparisons of the OPT models and LLaMA models with 4-bit quantization on WikiText-2. We set the group size as the length of rows for OPT models and 128 for LLaMA models following baselines for fair comparisons.
}
\resizebox{\linewidth}{!}{
\begin{tabular}{l|c|ccccccc|c|cc|c}
\Xhline{3\arrayrulewidth}
\multirow{2}{*}{\textbf{Method}} & 
\multirow{2}{*}{\textbf{Bits}} & 
\multicolumn{7}{c|}{\textbf{OPT}} & 
\multicolumn{4}{c}{\textbf{LLaMA}}
\\
\cline{3-9} \cline{10-13} 
 & & \textbf{125M} & \textbf{350M} & \textbf{1.3B} & \textbf{2.7B} & \textbf{6.7B} & \textbf{13B} & \textbf{30B} & 
 \textbf{1-7B} &
 \textbf{2-7B} & \textbf{2-13B} & 
 \textbf{3-8B}\\
\Xhline{3\arrayrulewidth}

OPTQ~\cite{frantar2022optq} & 4 & 31.12 & 24.24 & 15.47 & 12.87 & 11.39 & 10.31 & 9.63 & 6.22  & 5.69 &  4.98 & 7.63\\
LUT-GEMM~\cite{park2022lut} & 4 & 31.93 & 24.09 & 16.15 & 13.34 & 12.09 & 10.40 & 9.99 & 5.94 & 5.78 & 5.06 & 6.85\\
AWQ~\cite{lin2023awq} & 4 & 31.66 & 7.4e3 (outlier) & 15.22 & 13.19 & 11.23 & - & - & 5.78 & 5.60 &  4.97 & - \\
\hline
\textbf{Ours (Acc.)} & 4 & \textbf{28.72} & \textbf{21.59} & \textbf{14.98} & \textbf{12.65} & \textbf{10.95} & \textbf{10.20} & \textbf{9.63} & \textbf{5.76}  & \textbf{5.58} & \textbf{4.96} & \textbf{6.46} \\
\Xhline{3\arrayrulewidth}
\end{tabular}
}
\label{tab:4bit}
\end{table}

We previously considered lower-bit quantization because we aim to push the accuracy-efficiency boundary to lower bits with minimal accuracy compromise. This is meaningful for large-scale LLMs, where even at 3 bits, they remain memory-bound. As analyzed using the Roofline model shown in Figure 5 of \cite{yuan2024llm}, for Nvidia A6000 GPUs, the turning point from memory-bound to compute-bound is 200 arithmetic intensity (OPs/bytes). For LLaMA-7B models, all the operators in the decode/generation phase have around or less than 1 arithmetic intensity, as shown in Table 1 of \cite{yuan2024llm}. Even at 4 bits, the arithmetic intensity is approximately 1 $\div$ 3 $\times$ 32 = 8 (same ops but $\nicefrac{4}{32}$ fewer memory accesses), which is far less than the turning point of 200 and thus remains memory-bound, let alone larger models like LLaMA-70B or beyond. Reducing from 4 bits to 2 bits can help increase the arithmetic intensity and thus the theoretically maximum performance by 2x, from 6144G OPS to 12288G OPS. If memory is not a bottleneck for much smaller cases or prefill stages, higher bits can be used for better accuracy. Our goal is to offer an additional option and trade-off for large, memory-bound cases, without forcing the exclusive use of 2 bits.

\section{\hry{Comparison with MSFP}}
\label{sec:msfp}

MSFP~\cite{darvish2020pushing} is an important prior work that employs a shared exponent across a group of elements and shifts the mantissa accordingly, mimicking multiplication by powers of two. In contrast, we clarify that our approach differs from MSFP in two key aspects:

\begin{enumerate}
    \item \textbf{Nature of Approach}: MSFP uses shared exponents but relies on various shifted mantissa to represent the weights; without this, all weights would collapse to the same value. In contrast, we do not use shared exponents for scaling factors and eliminate the need for mantissa. In particular, each scaling factor is represented as a distinct power-of-two integer (equivalent to the exponents in floating-point numbers, completely removing the mantissa bits). In this way, the multiplication between a floating-point activation and a power-of-two integer scaling factor can be simplified to adding the corresponding integer to the exponent bit of the floating-point activation, as described in Fig.~\ref{fig:shiftaddllm} (c). In addition, rather than sharing the exponents, the entire scaling factor in ShiftAddLLM is shared across groups of binary weights in a column/block-wise manner, as illustrated in Fig.~\ref{fig:block-wise} (a) and detailed in Sec.~\ref{sec:shiftaddllm-2}, carefully designed to optimize both weight quantization and output activation errors without conflicts. Hence, there are clear differences between the MSFP datatype and our quantization scheme. In fact, our method is orthogonal to MSFP and can be combined with it by representing input activations in MSFP for more aggressive performance improvements.

    \item \textbf{Determining Shared Exponents or Scaling Factors}: The method for determining shared exponents in MSFP or shared scaling factors in our quantization scheme is different. MSFP selects the maximum exponent to share across the bounding-box size, i.e., the number of elements sharing one exponent~\cite{darvish2020pushing}, which is simpler in implementation yet might not be as adaptive. In contrast, in our ShiftAddLLM, the reparameterized binary weights and scaling factors result from multi-objective optimization. This optimization adaptively designs scaling factor patterns to avoid conflicts between optimizing weight errors and optimizing output activation errors.
\end{enumerate}

Finally, in terms of the performance outcomes, MSFP at 4 bits (1-bit sign and 3-bit mantissa) already suffers from large quantization errors, as evidenced by the significant KL divergence shown in Fig. 3 of \cite{darvish2020pushing}. In contrast, our ShiftAddLLM at 3 or 4 bits can still achieve comparable accuracy to FP baselines.
To directly compare ShiftAddLLM with MSFP, we conducted additional experiments to compare (1) quantization errors and (2) KL divergence using both methods against their floating-point counterparts. We randomly selected ten weight matrices from OPT-350M, quantizing or reparameterizing them using both methods. The results, as summarized in Tab.~\ref{tab:msfp}, indicate that ShiftAddLLM consistently outperforms MSFP, achieving lower KL divergence by 0.0065, 0.0271, and 0.0952, and reducing quantization errors by 1707.3, 3251.1, and 5862.0 at 4-bit, 3-bit, and 2-bit quantization, respectively.

\begin{table}[h]
\centering
\setlength{\tabcolsep}{3.6pt}
\renewcommand{\arraystretch}{1.2}
\caption{Comparison between MSFP and ShiftAddLLM with varying bits on KL Divergence and Quantization Error.}
\resizebox{0.8\linewidth}{!}{
\begin{tabular}{l|c|c|c}
\Xhline{3\arrayrulewidth}
\textbf{Methods} & \textbf{Bits} & \textbf{Avg. KL Divergence} & \textbf{Avg. Quant. Error} \\
\Xhline{3\arrayrulewidth}
MSFP (bounding-box size = 128) & 4 & 0.0117 & 4129.1 \\
ShiftAddLLM (group size = 128) & 4 & 0.0052 & 2421.8 \\
MSFP (bounding-box size = 128) & 3 & 0.0434 & 7859.9 \\
ShiftAddLLM (group size = 128) & 3 & 0.0163 & 4608.8 \\
MSFP (bounding-box size = 128) & 2 & 0.1485 & 14355.7 \\
ShiftAddLLM (group size = 128) & 2 & 0.0533 & 8493.7 \\
\Xhline{3\arrayrulewidth}
\end{tabular}
}
\label{tab:msfp}
\end{table}

\section{\hry{Additional Clarifications on Eyeriss}}
\label{sec:asic}

As emphasized in Sec.~\ref{sec:exps}, our primary focus is on GPU acceleration, specifically through the development of dedicated CUDA kernel support. It is worth noting that, we intentionally did not delve into specific ASIC designs in the main manuscript, which were referenced only to demonstrate potential energy savings.

To clarify the Eyeriss in estimating the energy costs, Eyeriss~\cite{chen2016eyeriss} is a well-known energy-efficient reconfigurable accelerator architecture designed for deep convolutional neural networks (CNNs). It optimizes both dataflow and memory access to reduce energy consumption during neural network processing. In our work, we adapt the Eyeriss architecture by modifying its MAC (Multiply-Accumulate) array, a key component responsible for performing heavy arithmetic computations in CNNs. Instead of using traditional MAC units across the array, we replace selected units with shift, add, and lookup table (LUT) operations, aligning with our proposed ShiftAddLLM approach. This modification significantly reduces both the area and power requirements, with savings ranging from 26\% to 89\% in different configurations. 
We refer readers to Fig. 4 of NASA~\cite{shi2022nasa}, which visually demonstrates the design principles of the overall architecture, and illustrates how replacing traditional MAC units with shift and add operations leads to significant reductions in both area and energy consumption. By adapting these principles, we enhance Eyeriss to better align with the computational needs of both LLMs and ShiftAddLLMs while maintaining power and area efficiency.

\end{document}